%% file: main.tex
\documentclass[10pt,twocolumn,letterpaper]{article}

\usepackage[pagenumbers]{cvpr}
\usepackage{algorithm}
\usepackage{algorithmic}
\usepackage{booktabs} 
\usepackage{multirow}
\usepackage{amssymb}
\usepackage{pifont}
\newcommand{\cmark}{\ding{51}}%
\newcommand{\xmark}{\ding{55}}%
\usepackage{amsmath}
\usepackage{adjustbox}
\usepackage{float}

\usepackage[misc]{ifsym}
\definecolor{cvprblue}{rgb}{0.21,0.49,0.74}
\usepackage[pagebackref,breaklinks,colorlinks,allcolors=cvprblue]{hyperref}
\usepackage{makecell}
\usepackage{listings}
\usepackage{framed}
\usepackage{lipsum}
\usepackage{enumitem}
\usepackage{graphicx}
\usepackage{caption}
\usepackage{colortbl}
\usepackage{tabularx}
\usepackage{cancel}
\usepackage{pifont}
\definecolor{cvprblue}{rgb}{0.21,0.49,0.74}

\newcommand\blfootnote[1]{%
  \begingroup
  \renewcommand\thefootnote{}\footnote{#1}%
  \addtocounter{footnote}{-1}%
  \endgroup
}
\newcommand{\mycc}{\cellcolor{blue!20}}

\PassOptionsToPackage{table}{xcolor}

\title{NADER: Neural Architecture Design via Multi-Agent Collaboration}

\author{Zekang Yang$^{1}$ \quad Wang Zeng$^{1}$ \quad Sheng Jin$^{2,1}$ \textsuperscript{\Letter} \quad Chen Qian$^{1}$ \quad Ping Luo$^{2}$ \quad Wentao Liu$^{1}$ \textsuperscript{\Letter}  \\
$^{1}$ SenseTime Research and Tetras.AI \quad
$^{2}$ The University of Hong Kong \\
\tt\small \{yangzekang, zengwang, jinsheng, qianchen, liuwentao\}@tetras.ai  \quad pluo@cs.hku.hk}

\begin{document}
\maketitle

\input{sections/0_abstract}

\input{sections/1_intro}

\input{sections/2_related}

\input{sections/3_method}
\input{sections/4_experiment}
\input{sections/5_conclusion}

{
    \small
    \bibliographystyle{ieeenat_fullname}
    \bibliography{main}
}

\clearpage
\appendix
\setcounter{table}{0}
\renewcommand{\thetable}{A\arabic{table}}
\setcounter{figure}{0}
\renewcommand{\thefigure}{A\arabic{figure}}
\setcounter{section}{0}
\renewcommand{\thesection}{A\arabic{section}}
\input{sections/appendix}

\end{document}

%% file: sections/0_abstract.tex
\begin{abstract}
Designing effective neural architectures poses a significant challenge in deep learning. While Neural Architecture Search (NAS) automates the search for optimal architectures, existing methods are often constrained by predetermined search spaces and may miss critical neural architectures. In this paper, we introduce NADER (Neural Architecture Design via multi-agEnt collaboRation), a novel framework that formulates neural architecture design (NAD) as a LLM-based multi-agent collaboration problem.
NADER employs a team of specialized agents to enhance a base architecture through iterative modification.
Current LLM-based NAD methods typically operate independently, lacking the ability to learn from past experiences, which results in repeated mistakes and inefficient exploration. To address this issue, we propose the Reflector, which effectively learns from immediate feedback and long-term experiences. Additionally, unlike previous LLM-based methods that use code to represent neural architectures, we utilize a graph-based representation. This approach allows agents to focus on design aspects without being distracted by coding.
We demonstrate the effectiveness of NADER in discovering high-performing architectures beyond predetermined search spaces through extensive experiments on benchmark tasks, showcasing its advantages over state-of-the-art methods. The codes will be released soon.
\blfootnote{\Letter : Corresponding Authors.}
\end{abstract}

%% file: sections/1_intro.tex
\section{Introduction}
Recent advancements in Large Language Models (LLMs) have significantly impacted various fields thanks to their exceptional ability to integrate knowledge and follow instructions~\cite{achiam2023gpt,touvron2023llama}
Leveraging techniques such as chain-of-thought reasoning~\cite{wei2022chain}, Retrieval-Augmented Generation (RAG)~\cite{lewis2020retrieval}, and multi-agent collaboration frameworks~\cite{park2023generative}, LLMs show great potential in creative tasks, including software development~\cite{hong2023metagpt}, neural architecture search~\cite{zheng2023can}, hyperparameter optimization~\cite{zhang2023using,yang2024autommlab}, and scientific discovery~\cite{lu2024aiscientist}.
\begin{figure}
    \centering
    \includegraphics[width=0.49\textwidth]{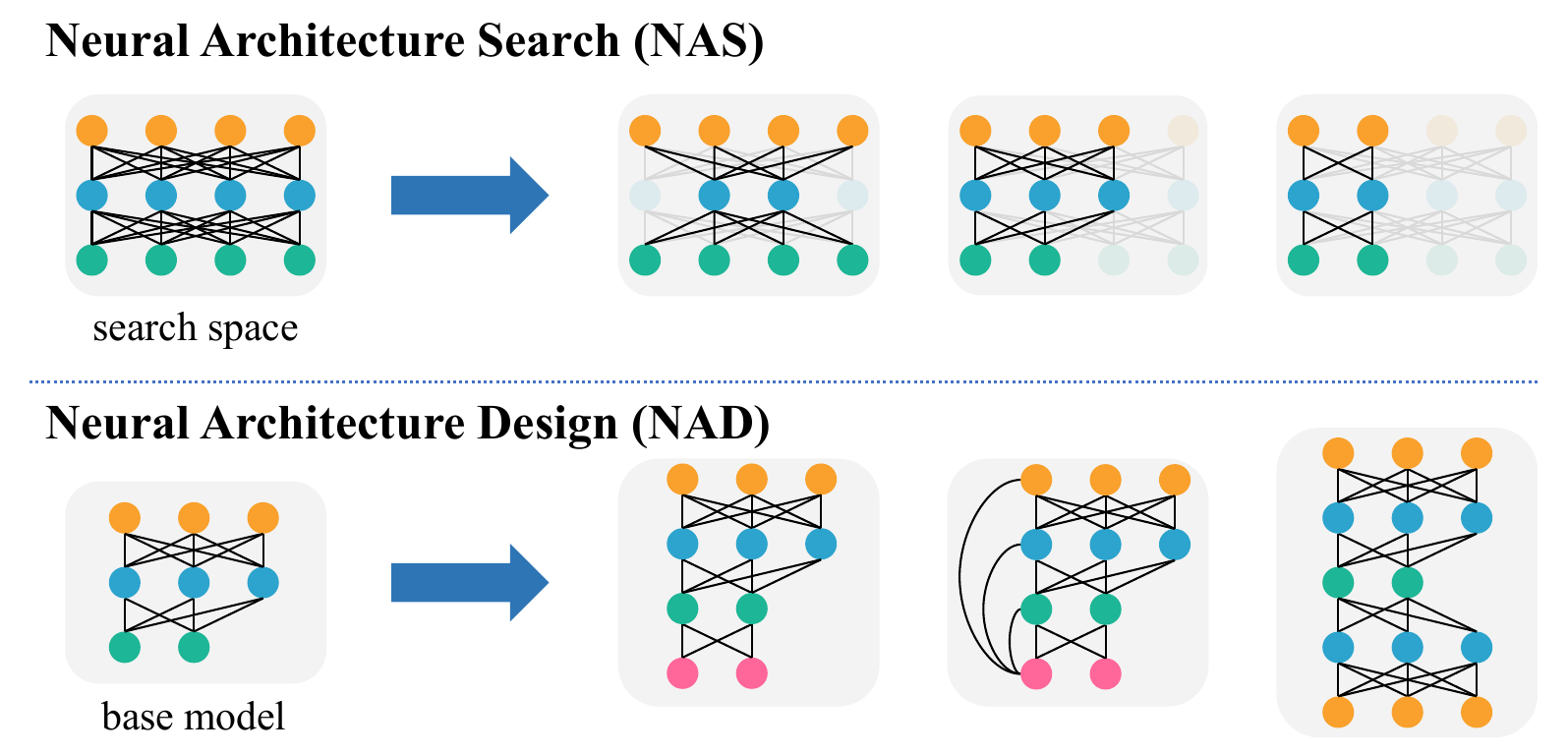}
    \caption{Traditional Neural Architecture Search (NAS) approaches aim to discover high-performing neural architectures within an expert-defined search space. In contrast, Neural Architecture Design (NAD) is not constrained by predefined search spaces, allowing for the exploration of entirely new architectures.
    }
    \label{fig:teaser}
\end{figure}
Designing neural network architectures that meet specific performance criteria (\eg, high accuracy or low latency) remains a significant challenge in deep learning. This process often involves extensive manual efforts and costly trial-and-error cycles. To address these challenges, existing studies have leveraged Neural Architecture Search (NAS) to automate the discovery of high-performing architectures, achieving notable results across diverse domains~\cite{zoph2016neural,liu2018darts}. However, traditional NAS methods typically rely on expert-defined search spaces, limiting their capacity to uncover truly novel architectures.

Motivated by advancements in large language models (LLMs), recent research has explored a new direction in Neural Architecture Design (NAD) driven by LLMs, moving beyond conventional predetermined search spaces. Unlike traditional NAS approaches, NAD fosters innovation by enabling the exploration of entirely new architectures within an open architecture space. However, this endeavor presents several challenges.
First, many existing studies rely on evolutionary computation or reinforcement learning for architecture modifications, which often result in inefficient, undirected exploration, leading to training and evaluation of an overwhelming number of neural network architectures.  
Second, existing LLM-based methods perform NAD tasks independently, without benefiting from immediate feedback or past experiences, leading to repeated mistakes and unnecessary trail-and-error processes.
Additionally, code-generating LLMs often struggle to maintain focus on neural architecture design, becoming sidetracked by non-essential aspects like text style and code syntax.
Such distractions can significantly diminish the efficiency of the automatic exploration.

To tackle these challenges, we propose NADER (Neural Architecture Design via multi-Agent collaboRation), an LLM-driven NAD method. NADER begins with a well-established network, such as ResNet~\cite{he2016deep}, and seeks to enhance it through iterative modifications. 
The framework comprises a research team and a development team, each consisting of multiple collaborating agents. This multi-agent approach facilitates more efficient and directed exploration. 
The research team is responsible for proposing improvement proposals based on insights drawn from recent academic literature and model optimization trees.
Each proposal contains a model to be improved and a modification suggestion.
Subsequently, the development team implements these proposals and evaluate the performance of the modified network.
The performance is then relayed back to the research team, informing subsequent proposals and fostering a continuous cycle of improvement.

``Error is often the correct guide." We introduce the Reflector, which effectively incorporates immediate feedback from the external environment and leverages historical experiences.
Reflector tracks the behavior of Modifier and reflect on it to gain experience.
Given the proposed modification suggestions, the Reflector queries the experience pool to retrieve relevant experience, providing the Modifier with contextual information for implementation.
This reduces the likelihood of repeating past mistakes and increases the success rate of a good modifications.

Furthermore, to help the development team concentrate on network structure, we propose a graph-based network representation. Instead of generating executable code directly, NADER represents neural architectures as single directed acyclic graphs (DAGs), where nodes represent operations (\eg, convolution, normalization) and edges denote information flow. This graph representation allows NADER to prioritize high-level structural decisions, avoiding distractions from low-level code implementation details.

We demonstrate the effectiveness of NADER in discovering high-performing neural architectures that extend beyond predetermined search spaces. Through extensive experiments, we showcase the advantages of NADER over state-of-the-art NAS and NAD methods. Our contributions are summarized as follows:
\begin{itemize}
    \item We introduce NADER, a novel LLM-driven framework for neural architecture design that utilizes a multi-agent collaboration strategy to automatically explore novel neural architectures beyond predetermined search spaces. 
    \item  We introduce Reflector, which effectively incorporates immediate feedback from the external environment and experiences reflected from historical trajectories, thereby reducing repeated mistakes and enhancing efficiency.
    \item We introduce a graph-based representation for neural network architectures, facilitating a focus on high-level structural decisions while minimizing distractions from low-level coding details.
\end{itemize}

%% file: sections/2_related.tex
\section{Related Work}
\subsection{Neural Architecture Search (NAS)}
NAS aims to automatically discover optimal neural network architectures from predefined search spaces. NAS methods have achieved impressive success across various tasks. 
Early approaches relied on reinforcement learning~\cite{williams1992simple,pham2018efficient}, evolutionary strategies~\cite{real2019regularized} and Bayesian optimisation~\cite{kandasamy2018neural} to explore the search space.
More recent methods have shifted towards weight sharing techniques. These approaches~\cite{liu2018darts,dong2019one,hu2020dsnas,xu2019pc,xie2018snas,zhang2021idarts,dong2019searching,chen2020drnas,ye2022b} train a single super-network, enabling efficient exploration of various sub-networks (architectures) within the search space.
However, these methods often rely on predetermined search spaces, which limits their ability to discover truly novel architectures. In contrast, our approach addresses the Neural Architecture Design (NAD) task, which is not restricted by predetermined search spaces.

\subsection{LLM Agents}
As large language models have demonstrated impressive
emergent capabilities and have gained immense popularity, recent research has focused on LLM-based autonomous agents to solve complex problems~\cite{xi2023rise}. 
These agents utilize the foundational capabilities of LLMs, displaying remarkable skills in memory~\cite{park2023generative}, planning~\cite{liu2023bolaa}, tool use~\cite{schick2024toolformer,qin2023toolllm}, and reflection~\cite{shinn2024reflexion,zhao2024expel}.
Systems like Visual ChatGPT~\cite{wu2023visual} and HuggingGPT~\cite{shen2024hugginggpt} employ LLMs to orchestrate diverse AI models for complex tasks. Visual programming approaches like~\cite{gupta2023visual} utilize LLMs to generate programs for solving visual tasks based on natural language instructions. 
There are also studies on multi-agent communication~\cite{liang2023encouraging} and collaboration~\cite{hong2023metagpt,qian2023communicative,chen2023agentverse}. For example, MetaGPT~\cite{hong2023metagpt} utilizes an assembly line paradigm, assigning roles to LLM-based agents for collaborative problem-solving.
Our work extends this concept by introducing a team of specialized LLM agents for neural architecture design. 
\begin{figure*}[t]
    \centering
    \includegraphics[width=0.99\textwidth]{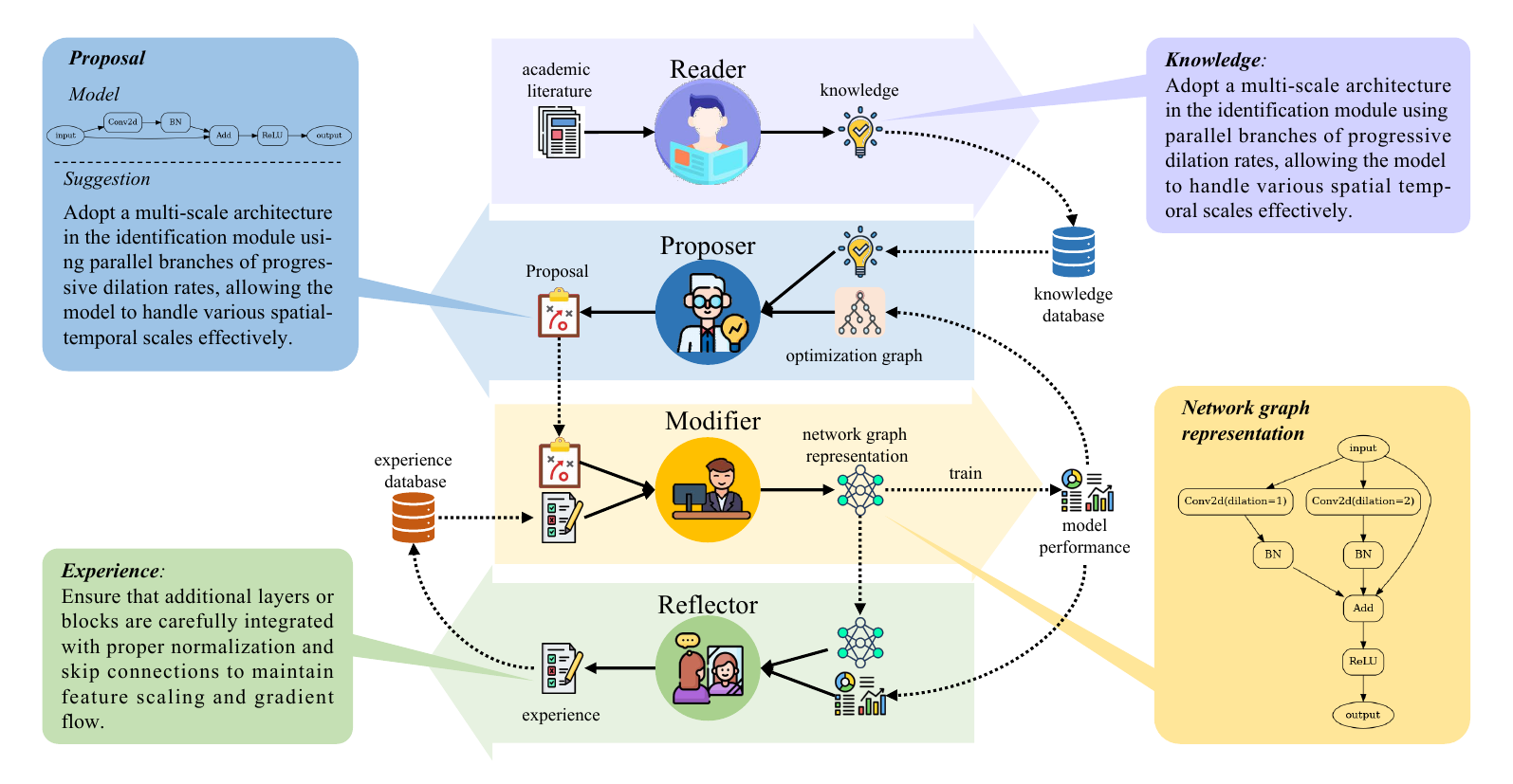}
    \caption{Overview of NADER framework. The Reader continuously learns from academic literature, while the Proposer identifies the most promising candidate networks and suggests modifications. The Modifier implements these suggestions, and the Reflector analyzes and provides feedback on the results. The performance of the modified network is relayed back to the Proposer, informing subsequent proposals and fostering a continuous cycle of improvement.}
    \label{fig:framework}
\end{figure*}
\subsection{LLMs for NAS}
Emerging research explores the use of LLMs for NAS. 
GENIUS~\cite{zheng2023can} is a LLM-based NAS algorithm that adopts a straightforward prompt-based search process that prompts GPT-4 to propose designs from a given search space with a handful of examples.
EvoPrompting~\cite{chen2024evoprompting} utilizes code-LLMs as mutation and crossover operators for an evolutionary neural architecture search (NAS) algorithm. 
LLMatic~\cite{nasir2023llmatic} extends EvoPrompting by introducing quality diversity algorithms to discover diverse and robust solutions.
Our work differentiates them by introducing a multi-agent framework, allowing for a more sophisticated and collaborative exploration of the search space.

%% file: sections/3_method.tex
\section{Neural Architecture Design via Multi-Agent Collaboration}
\subsection{Overview}
We present the Neural Architecture Design via Multi-Agent Collaboration (NADER) system, a novel approach for neural architecture design that harnesses the capabilities of large language models (LLMs) and multi-agent collaboration. This system begins with a base network and iteratively refines its structure for optimal performance.

As illustrated in Figure~\ref{fig:framework}, NADER is composed of a Research Team (Reader and Proposer) and a Development Team (Modifier and Reflector). The Reader is responsible for extracting knowledge from the latest academic literature, ensuring that the system stays current with recent advancements. The Proposer utilizes the knowledge provided by the Reader to generate modification proposals, which include a candidate network and a corresponding modification suggestion. The Modifier practically implements these proposals to get new networks and their corresponding performance metrics.
Algorithm \ref{alg:nader} shows the design process of neural architecture
The Reflector tracks and reflects on Modifier's logs and offers guidance to Modifier.

\subsection{Graph-based Neural Architecture Representation}
The initial step in Neural Architecture Design (NAD) involves defining a representation for network architectures.
While LLMs excel at understanding and generating program code, this can be suboptimal for NAD scenarios. 
Program code often contains extraneous information, such as text style and formatting, which can distract LLMs from focusing on the network structure, thereby reducing exploration efficiency.
\begin{figure}[t]
    \centering
    \includegraphics[width=0.5\textwidth]{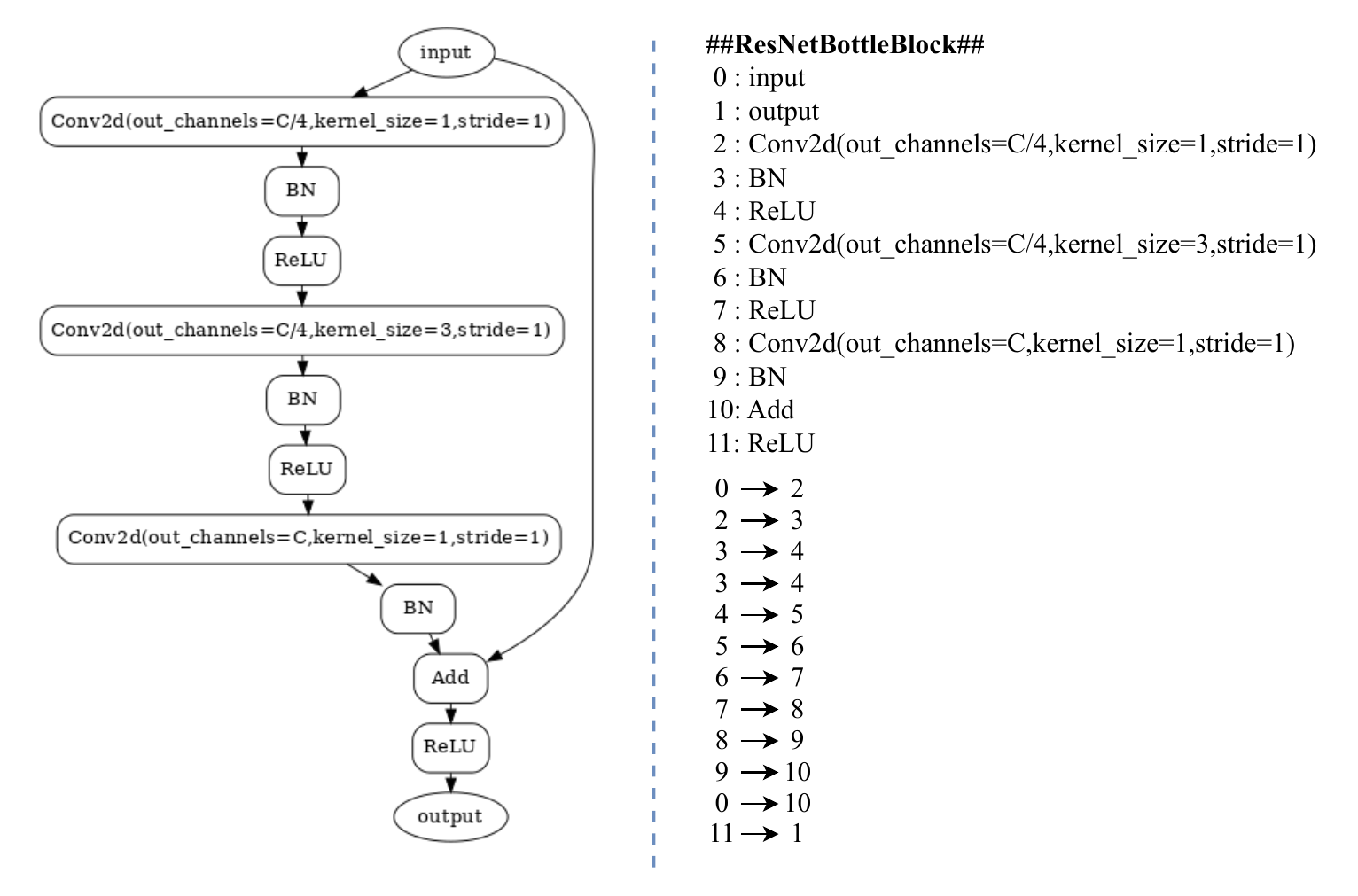}
    \caption{Illustration of graph-based neural architecture representation. \textbf{Left}: Visualization of DAG. \textbf{Right}: Text representation of DAG for LLM understanding.}
    \label{fig:graph}
\end{figure}
To address this issue, we introduce a graph-based representation for network structures. As shown in Figure~\ref{fig:graph}, each neural network architecture can be represented as a single directed acyclic graph (DAG). In this graph, nodes correspond to input, output and fundamental operators within a neural network (\eg, convolutional layer, matrix addition), and an edge from operator $N_i$ to operator $N_j$ indicates that $N_j$ receives the output of $N_i$ as input. 

This graph representation eliminates irrelevant factors, allowing LLMs to concentrate solely on network structure design without being bogged down by low-level implementation details. 
Meanwhile, the graph representation can easily and accurately check whether different neural architectures are isomorphic, which can help organize neural architecture optimization paths more effectively.
Furthermore, the graph representation is more compact and efficient, as it can be represented with fewer tokens.
In addition, such a representation is more readable and easier to understand. 
In NADER, this graph representation is utilized for both the input and output of the agents.

\subsection{Research Team}
The Research Team is responsible for providing high-level network improvement proposals, a crucial aspect of NAD. The team is comprised of a Reader agent and a Proposer agent. 
The Reader agent distills knowledge from academic literature, ensuring the system is informed by the latest developments. The Proposer agent then leverages this knowledge to generate high-quality improvement proposals.

\subsubsection{Reader}
Staying current with the latest academic developments is crucial for neural architecture design. However, LLMs often lag behind the latest knowledge and struggle to absorb new information, hindering their ability to propose novel and valuable improvements. To overcome this challenge, we propose the Reader agent for literature-based knowledge discovery, Algorithm \ref{alg:reader} illustrates this process.
The Reader agent analyzes recent academic papers on neural architectures and related advancements. It automatically downloads recent articles from websites, evaluates their relevance to neural architecture design, and extracts key knowledge and novel ideas that could potentially improve neural network performance. These knowledge are then archived in a database for future use.
\textbf{Revelant Papers Selection.} A large number of crawled papers may not be relevant to neural architecture design. Reader quickly find potentially useful papers by reading paper abstracts.
\textbf{Knowledge Extraction.} The content of a paper is usually lengthy. By focusing on the method section of the paper, Reader can better extract the core innovation of the paper and summarize the inspiration that may be useful for neural architecture design.
\begin{algorithm}[t]
    \caption{Knowledge extraction process of Reader}
    \begin{algorithmic}[1]
    \label{alg:reader}
    \REQUIRE{Papers $\{P_i\}_{i=1}^N$ and their abstracts $\{A_i\}_{i=1}^N$; Knowledge vector database $\mathcal{K}$; A pre-trained LLM; Text to embedding model $\mathcal{M}$.}
    \ENSURE{Updated $\mathcal{D}_{insp}$.}
    \STATE Initialize the agent $\mathbf{Reader}$ with the LLM.
    \FOR{$i\leftarrow 1$ \TO $N$}
        \IF{$\mathbf{Reader}$ judge the $P_i$ is relevant based on $A_i$}
            \STATE $ks\leftarrow \mathbf{Reader}$ extracts knowledge from $P_i$.\FOR{each element $k$ in $ks$}
                \STATE $embed\leftarrow$ $\mathcal{M}$ extracts the embedding of $k$.
                \STATE $\mathcal{K}\leftarrow$ Update $\mathcal{K}$ with $embed$ and $k$.
            \ENDFOR
        \ENDIF
    \ENDFOR
    \end{algorithmic}
\end{algorithm}
\subsubsection{Proposer}
The Proposer agent is responsible for guiding the architecture modification process.
By strategically navigating the model modification graph and selecting the most promising candidate network and proposing modification suggestions, the Proposer agent optimizes the overall architecture design process, maximizing the potential for performance improvements while minimizing the risk of exploring unproductive paths.
\textbf{Network Modification Tree.} 
Network modification tree records the history of modification during NAD. The tree's root represents the basic network, with each node representing a modified network architecture and each edge representing an applied modification suggestion. The performance of each node is recorded within the graph.
\textbf{Candidate Network Selection.} Given the network modification graph, the Proposer selects the most promising node as the candidate network for the next modification. The Proposer leverages the performance data recorded within the graph to guide its selection. It prioritizes nodes with higher performance scores, indicating successful past modifications. 
Specifically, it explores the network modification graph using a combination of depth-first search (DFS) and breadth-first search (BFS). DFS allows for a deeper exploration of promising paths within the graph, while BFS ensures a wider coverage of the search space.
\textbf{Modification Suggestion Selection.} 
The knowledge pool constructed by the Reader are abundant and vary in quality. The Proposer selects a set of promising knowledge from the database as the suggestions to be applied to the chosen candidate network. 
It uses reflections from previous experiences to retrieve modification suggestions at different similarity levels, balancing exploration of novel ideas with the exploitation of past successes. It prioritizes suggestions that have proven effective in similar contexts or that address common error patterns identified during past modifications.

\begin{algorithm}[t]
    \caption{Neural architecture design process of NADER}
    \begin{algorithmic}[1]
    \label{alg:nader}
    \REQUIRE{The initial architecture $arch_0$; A pre-trained LLM; Knowledge database $\mathcal{K}$; Design experience database $\mathcal{E}$; Number of architectures to generate $N$; Max number of retry after failed $max\_retry$.}
    \ENSURE{Best architecture $arch_{best}$; Historical records $\mathcal{H}$; Network modification tree $\mathcal{T}$.}
    \STATE Initialize the $\mathbf{Proposer}$ and $\mathbf{Modifier}$ with LLM.
    \STATE Initialize model optimization tree $\mathcal{T}$ with $arch_0$.
    \STATE $n\leftarrow 0$ Initialize the number of architectures generated.\WHILE{$n < N$}
        \STATE $insps\leftarrow$ Retrieve inspirations from $\mathcal{K}$.
        \STATE $prop\leftarrow \mathbf{Proposer}$ generate a proposal based on $insps$ and $\mathcal{T}$.
        \STATE $e\leftarrow$ Retrieve experiences from $\mathcal{E}$ by query $prop$.\STATE $conv\leftarrow$ Initialize conversation with $prop$ and $e$.
        \FOR{$i\leftarrow1$ \TO $max\_retry$}
            \STATE $arch_n\leftarrow \mathbf{Modifier}$ generate a new architecture according to $conv$.
            \STATE $error\leftarrow$ Check the correctness of $arch_n$.
            \STATE $conv\leftarrow$ Update $conv$ with $arch_n$ and $error$.
            \STATE \textbf{if} $error$ is empty \textbf{then} \textbf{break}
        \ENDFOR
        \IF{$error$ is empty}
            \STATE $arch\_acc_{n}\leftarrow$ Train architecture $arch_n$ on train dataset and get its test accuracy on test dataset.
            \STATE $\mathcal{T}\leftarrow$ Update $\mathcal{T}$ with $prop$, $arch_n$ and $arch\_acc_n$.\STATE $n\leftarrow n+1$
        \ENDIF
        \STATE $\mathcal{H}\leftarrow$ Update $\mathcal{H}$ with $prop$, $conv$ and $arch\_acc_n$.
    \ENDWHILE
    \STATE $arch_{best}\leftarrow$ Get the best architecture from $\{arch_{i}\}_{i=1}^N$.
    \end{algorithmic}
\end{algorithm}
\subsection{Development Team}
The Development Team is responsible for implementing the architectural modifications suggested by the Research Team. Its goal is to make reasonable and accurate changes to the neural architecture. The team consists of the Modifier and the Reflector, which collaborate through a multi-round dialogue to refine and correct the design.

\subsubsection{Modifier}
The Modifier agent modifies the given architecture based on the modification suggestions. It performs modifications hierarchically, starting from a single block and progressing to the entire network.
Initially, it implements a new cell block using the candidate block and modification suggestions. This process considers both macro-level guidance (\eg., introducing multi-scale information) and micro-level design experience (\eg, residual connections before output).
Given the new cell block, it then implements corresponding and compatible stem and downsample blocks. The prompt for this step includes: (1) the cell block structure, (2) examples cell blocks and their corresponding stem blocks, and (3) design principles (\eg, downsampling by at least 2x in the stem to prevent memory overflow).
\textbf{Isomorphic Graph Checking.} When the newly generated neural architecture is detected to be isomorphic to the existing neural architecture, no repeated training is performed.
\textbf{Graph-to-Code Conversion.} NADER adopts the graph representation for neural network architectures. Once a valid computational graph is designed, this component translates the graph representation into executable code.
\begin{algorithm}[t]
    \caption{Design experience extraction process}
    \begin{algorithmic}[1]
    \label{alg:reflector}
    \REQUIRE{Historical records $\mathcal{H}$; Design experience vector database $\mathcal{E}$; Network modification tree $\mathcal{T}$; A pre-trained LLM; Text to embedding model $\mathcal{M}$.}
    \ENSURE{Updated $\mathcal{E}$.}
    \STATE Initialize the agent $\mathbf{Reflector}$ with LLM.
    \FOR{each $h_i$ in $\mathcal{H}$}
        \STATE $p_i, arch_i,arch\_acc_i,error_i\leftarrow$ Get proposal, architecture, test accuracy and error information from $h_i$.\STATE $arch_{j}$, $arch\_acc_{j}\leftarrow$ Get the father architecture and its test accuracy of $arch_i$ from $\mathcal{T}$.
        \STATE $e_i\leftarrow \mathbf{Reflector}$ reflects on ($arch_{i}$, $arch\_acc_{i}$, $arch_{j}$, $arch\_acc_{j}$, $error_i$).
        \STATE $pe_i\leftarrow$ $\mathcal{M}$ extracts the embedding of $p_i$.
        \STATE $\mathcal{E}\leftarrow$ Update $\mathcal{E}$ with key $pe_i$ and value $e_i$.
    \ENDFOR
    \end{algorithmic}
\end{algorithm}
\subsubsection{Reflector} 
The Reflector agent tracks, analyzes and reflects on the results of the Modifier to support the modification process.
\textbf{Learn from Immediate Feedback (LIF).} 
We have developed a computational graph flow verification tool to ensure the generated neural architecture is valid and executable. It checks the connectivity and information flow within the computational graph, verifying that all operations are properly connected and that the overall structure is coherent. Given the neural architecture implemented by the Modifier, the Reflector utilizes this verification tool to analyze its executability. If an invalid structure is generated, the agent communicates with the Modifier to retry and correct the errors.
\textbf{Learn from Design Experience (LDE).} The Reflector also plays a crucial role in learning from historical records and summarizing design experience to guide future design process, Algorithm ~\ref{alg:reflector} demonstrates the process by which Reflector extracts design experience from historical records. 
It tracks the Modifier's historical records, including both successes and failures.
(1) Failure Records: This category stores records of modifications that led to a decrease in performance or resulted in an invalid architecture.
(2) Failure-to-Success Records: This category stores records of modifications that initially failed but were later corrected. 
(3) Success Records: This category stores records of successful modifications that led to an improvement in performance. 
The Reflector analyzes the error causes associated with each failure record, generating design experience and categorizing them based on the type of error (\eg, invalid connection, memory overflow, performance degradation). This analysis helps identify common error patterns and develop strategies for avoiding them in future design.
To facilitate efficient retrieval of relevant experience, the Reflector utilizes a database where queries are modification suggestions, and keys are lists of error/success experience for base, stem, and downsample blocks. For each modification trial, the Reflector queries the database using the proposed modification suggestion as input. It then retrieves 5 related design experience from the database, based on the similarity between the current suggestion and past records. The Reflector then provides these 5 relevant experience as context to the Modifier. The Modifier leverages the context to make more informed decisions during the design process, reducing the likelihood of repeating past errors and increasing the success rate of the design.

%% file: sections/4_experiment.tex
\section{Experiment and Analysis}
\begin{table*}[t]
\centering
\caption{Comparisons with state-of-the-art NAS and NAD methods. NAS is limited by the expert-defined search space, \ie NAS-Bench-201, while NAD enables the exploration of entirely new architectures beyond the search space. We compare with $\diamondsuit$ methods with parameter sharing, $\heartsuit$ methods without parameter sharing, $\bigstar$ LLM-based methods. We report the mean validation and test accuracies for 5 three runs.
}
\label{tab:nasbench}
\scalebox{0.9}{\begin{tabular}{c|l|c|cc|cc|cc}
\toprule
 & \multirow{2}{*}{Method} & Search & \multicolumn{2}{c|}{CIFAR-10} & \multicolumn{2}{c|}{CIFAR-100} & \multicolumn{2}{c}{ImageNet16-120} \\
 & &  (archs) & validation & test & validation & test & validation & test \\ \midrule
\parbox{2.5mm}{\multirow{16}{*}{\rotatebox[origin=c]{90}{\textbf{NAS}}}} 
 & $\diamondsuit$ ENAS~\cite{pham2018efficient} & - & 37.51±3.19 & 53.89±0.58 & 13.37±2.35 & 13.96±2.33 & 15.06±1.95 & 14.84±2.10 \\
 & $\diamondsuit$ DARTS \cite{liu2018darts}  & - & 39.77±0.00 &  54.30±0.00 &  38.57±0.00 & 15.61±0.00 &  18.87±0.00 &  16.32±0.00 \\
 & $\diamondsuit$ SETN~\cite{dong2019one} & - & 84.04±0.28 & 87.64±0.00 & 58.86±0.06 & 59.05±0.24 & 33.06±0.02 & 32.52±0.21  \\
 & $\diamondsuit$ DSNAS \cite{hu2020dsnas}  & - & 89.66±0.29 & 93.08±0.13 &  30.87±16.40 & 31.01±16.38 & 40.61±0.09 & 41.07±0.09 \\
 & $\diamondsuit$ PC-DARTS \cite{xu2019pc} & -  & 89.96±0.15 & 93.41±0.30 & 67.12±0.39 & 67.48±0.89 & 40.83±0.08 & 41.31±0.22 \\
 & $\diamondsuit$ SNAS \cite{xie2018snas} & - & 90.10±1.04 & 92.77±0.83 & 69.69±2.39 & 69.34±1.98 & 42.84±1.79 & 43.16±2.64 \\ 
 & $\diamondsuit$ iDARTS \cite{zhang2021idarts} & - & 89.86±0.60 & 93.58±0.32 &  70.57±0.24 & 70.83±0.48 &  40.38±0.59 & 40.89±0.68 \\
 & $\diamondsuit$ GDAS \cite{dong2019searching} & - & 89.89±0.08 & 93.61±0.09 & 71.34±0.04 & 70.70±0.30 & 41.59±1.33 &  41.71±0.98  \\
 & $\diamondsuit$ DRNAS \cite{chen2020drnas} & - & {91.55±0.00} & 94.36±0.00  &  73.49±0.00 & 73.51±0.00 & 46.37±0.00 & 46.34±0.00  \\
 & $\diamondsuit$ $\beta$-DARTS \cite{ye2022b} & -  & {91.55±0.00} & 94.36±0.00  & 73.49±0.00 & 73.51±0.00 & 46.37±0.00 & 46.34±0.00  \\
 & $\diamondsuit$ $\Lambda$-DARTS \cite{movahedi2022lambda} & - & {91.55±0.00} & 94.36±0.00  &  73.49±0.00 & 73.51±0.00 & 46.37±0.00 & 46.34±0.00  \\ 
 & $\heartsuit$ REA~\cite{real2019regularized} & 500 &91.19±0.31 & 93.92±0.30 & 71.81±1.12 & 71.84±0.99 & 45.15±0.89 & 45.54±1.03 \\
 & $\heartsuit$ RS~\cite{bergstra2012random} & 500 & 90.93±0.36  & 93.70±0.36& 70.93±1.09 & 71.04±1.07& 44.45±1.10 & 44.57±1.25 \\
 & $\heartsuit$ REINFORCE~\cite{williams1992simple} & 500 & 91.09±0.37 & 93.85±0.37 & 71.61±1.12 & 71.71±1.09 & 45.05±1.02 & 45.24±1.18 \\
 & $\heartsuit$ BOHB~\cite{falkner2018bohb} & 500 & 90.82±0.53 & 93.61±0.52 & 70.74±1.29 & 70.85±1.28 & 44.26±1.36 & 44.42±1.49 \\
 & $\bigstar$GENIUS~\cite{zheng2023can} & 10 & 91.07±0.20 & 93.79±0.09 & 70.96±0.33 & 70.91±0.72 & 45.29±0.81 & 44.96±1.02 \\ 
 & $\bigstar$LLMatic~\cite{nasir2023llmatic} & 2000 & - & 94.26±0.13 & - & 71.62±1.73 & - & 45.87±0.96 \\
 & \mycc  ~~~~\textit{Optimal} & \mycc  - & \mycc 91.61 & \mycc 94.37 & \mycc 73.49 & \mycc 73.51 & \mycc 46.77 & \mycc 47.31 \\ \midrule
 \parbox{2.5mm}{\multirow{8}{*}{\rotatebox[origin=c]{90}{\textbf{NAD}}}}  
 & $\bigstar$LeMo-NADe-Gemini~\cite{rahman2024lemo} & 30 & 82.94 & 81.76 & 52.12 & 52.96 & 30.34 & 31.02 \\
 & $\bigstar$LeMo-NADe-GPT4~\cite{rahman2024lemo} & 30 & 90.90 & 89.41 & 63.38 & 67.90 & 27.05 & 27.70 \\
\cline{2-9}
 & \multirow{3}{*}{$\bigstar$ NADER (Random)} & 0 & 87.67±2.88 & 90.36±2.94 & 64.89±4.94 & 64.81±5.13 & 36.56±6.60 & 36.51±7.11 \\
 &  & 5 & 90.91±0.46 & 94.20±0.38 & 74.11±0.25 & 74.02±0.33 & 48.73±0.19 & 48.72±0.14 \\
 &  & 10 & 91.16±0.36 & 94.40±0.23 & 74.41±0.34 & 74.51±0.16 & \underline{50.07±0.75} & \underline{49.63±0.80} \\
\cline{2-9}
 & \multirow{4}{*}{$\bigstar$ NADER (ResNet)} & 0 & 90.83 & 93.97 & 70.42 & 70.86 & 44.53 & 43.63 \\
 &  & 5 & 91.17±0.24 & 94.52±0.22 & 73.29±1.86 & 73.12±1.09 & 47.98±0.73 & 47.99±0.38 \\
 &  & 10 & 91.18±0.23 & \underline{94.52±0.22} & \underline{74.71±0.45} & \underline{74.65±0.33} & 48.56±0.83 & 48.61±0.76 \\
 &  & 500 &  \textbf{91.55} & \textbf{94.62} & \textbf{75.72} & \textbf{76.00} & \textbf{50.20} & \textbf{50.52} \\
 \bottomrule
 \end{tabular}}
\end{table*}
\subsection{NAD Benchmark}
To facilitate developing and evaluating LLM-based Neural Architecture Design (NAD), we create a comprehensive benchmark dataset. 
We established five types of basic blocks as our starting points: (1) ResNet base block, (2) ResNet bottleneck block, (3) GoogLeNet block, (4) ConvNeXt block and (5) SE-ResNet block. For each basic block, we used GPT-4o to generate some initial macro-level and micro-level modification suggestions. 
We manually inspected and modified the generated modification suggestions, resulting in 10 high-quality macro-level and 10 micro-level modification suggestions for each basic block. There are $(10+10) \times 5 = 100$ test samples in total.
This allows us to evaluate the system's performance on a diverse set of architectural modifications, ranging from basic blocks to more complex structures inspired by state-of-the-art models. Macro-level modification suggestions don't have definitive answers and are more open-ended, while the micro-level modification suggestions have more specific, ground-truth answers. 
For each pair of basic block and its corresponding modification suggestions, we manually checked and generated ground-truth resulting blocks for reference. It is ensured that the resulting blocks reflect the modification suggestions and were executable.

\textbf{Evaluation Metrics.} To assess the performance of our approach, we use the 100 test samples from the NAD benchmark to evaluate the ability to generate valid and executable architectures that adhere to both macro-level and micro-level modification suggestions.
\begin{itemize}
    \item \textbf{Executability (E):} This metric quantifies the agents’ ability to generate executable code as the percentage of successfully executable modified architectures.
    \item \textbf{Quality (Q):} This metric assesses the ability of agents to generate a neural architecture that meets the modification suggestions, quantified as the percentage of modified architectures that successfully implement the specified modification suggestions among the executable ones.
    \item \textbf{Success Rate (SR):} The success rate is a holistic metric that considers both executability and quality, providing a comprehensive measure of the approach's effectiveness.
    \item \textbf{Number of Tokens (\# Tokens):} This metric represents the cost per execution, with a higher number of generated tokens indicating increased computational expense.
\end{itemize}
\subsection{Comparison with State-of-the-art Models}
\textbf{Experimental Setup.}
We conduct experiments to compare both NAS and NAD methods using the well-known NAS-Bench-201~\cite{dong2020bench} benchmark, covering evaluations on CIFAR10, CIFAR100~\cite{krizhevsky2009learning}, and ImageNet16-120~\cite{chrabaszcz2017downsampled}. The results are demonstrated in Table~\ref{tab:nasbench}. 
Our NADER accepts any network as the initial network. In our experiments, we report the results of utilizing ResNet~\cite{he2016deep} as the initial network and the results of using randomly selected networks from the NAS-Bench-201 search space. We report results for designing 5 and 10 new architectures. Each result reflects the mean and standard deviation from three repeated runs with different random seed.

For NAD, the newly designed architecture can be out of the scope of the search space.
To ensure fairness, (1) we constrain the architectures generated by NADER to have the same macro skeleton with NAS-Bench-201. (2) We constrain the FLOPs and parameter count of NAD methods not to exceed the largest network within the NAS search space: paramerts no more than 1.5M for all datasets, FLOPs no more than 0.2 GFLOPs for CIFAR10 and CIFAR100, and FLOPs no more than 0.05 GFLOPs for ImageNet16-120. (3) Model training adhered strictly to the NAS-Bench-201 settings, maintaining the original train/validation/test splits without altering any hyperparameters.

\textbf{Comparisons with NAS methods.}
NAS methods operate within a predetermined search space, focusing primarily on designing cell blocks for neural architectures. Given five operation choices and four nodes for each cell, the total search space expands to $5^6 = 15625$ candidated cells. We evaluated several prominent NAS algorithms, categorized as follows:
(1) Random search algorithms: random search (RS)~\cite{bergstra2012random} and random search with parameter sharing (RSPS)~\cite{li2020random};
(2) ES methods: REA~\cite{real2019regularized};
(3) RL algorithms: REINFORCE~\cite{williams1992simple} and ENAS~\cite{pham2018efficient};
(4) HPO methods: BOHB~\cite{falkner2018bohb}; 
(5) Differentiable algorithms: DARTS \cite{liu2018darts}, SETN~\cite{dong2019one}, DSNAS \cite{hu2020dsnas}, PC-DARTS \cite{xu2019pc}, SNAS \cite{xie2018snas}, iDARTS \cite{zhang2021idarts}, GDAS \cite{dong2019searching}, DRNAS \cite{chen2020drnas}, $\beta$-DARTS \cite{ye2022b}, $\Lambda$-DARTS;
(6) LLM-based methods: GENIUS~\cite{zheng2023can} and LLMatic~\cite{nasir2023llmatic}. We provide the optimal accuracy for reference (\textit{Optimal}), which is the maximum accuracy that can be achieved in the NAS-bench-201 search space. 
Our findings indicate that the NADER framework is robust; even when initialized with a random network, it can be effectively optimized after just a few iterations. Notably, in nearly all scenarios, NADER consistently designs architectures that surpass the optimal accuracy achievable within the predetermined search space, demonstrating the advantages of NAD over traditional NAS methods.

\textbf{Comparisons with NAD methods.} We also compare NADER against the recent NAD method, LeMo-NADe~\cite{rahman2024lemo}, which employs advanced LLMs like Gemini and GPT4-Turbo to refine neural networks based on expert-generated instructions. While LeMo-NADe shows promise, it still falls short compared to more recent NAS methods in certain respects. Our NADER approach significantly surpasses LeMo-NADe in both effectiveness and efficiency. For instance, with a random base network, NADER achieved 49.63\% accuracy on the ImageNet16-120 test set, compared to LeMo-NADe's 31.02\%. Furthermore, NADER requires only 10 trials—three times fewer than LeMo-NADe. These results highlight the potential of leveraging multi-agent collaboration for both effective and efficient NAD in an open architectural space.

\begin{table}[t]
    \centering
    \caption{Ablation studies of Research Team. Experiments are conducted on CIFAR10, CIFAR100, and ImageNet16-120 datasets. We report the results of neural architecture modifications for 5 iterations and show the mean and standard deviation from three repeated runs with different random seed.
    ``R'' and ``P'' represent Reader and Proposer respectively.
    }
    \label{tab:reader}
    \scalebox{0.61}{
    \begin{tabular}{cc|cc|cc|cc}
    \toprule
    \multirow{2}{*}{R} & \multirow{2}{*}{P}  & \multicolumn{2}{c|}{CIFAR-10} & \multicolumn{2}{c|}{CIFAR-100} & \multicolumn{2}{c}{ImageNet16-120} \\
     & & validation & test & validation & test & validation & test \\ 
    \midrule
    \xmark & \xmark & 90.37±0.65 & 94.01±0.42 & 72.55±0.27 & 72.49±0.40 & 47.12±1.26 & 47.14±1.30 \\
    \xmark & \cmark & 90.73±0.32 & 93.73±0.38 & 73.55±1.26 & 73.45±1.32 & 47.88±0.29 & 47.92±0.34 \\
    \cmark & \xmark & 90.86±0.19 & 91.72±2.52 & 73.67±0.85 & 73.37±0.73 & 47.89±1.16 & 47.91±1.00 \\
    \cmark & \cmark & \textbf{90.91±0.46} & \textbf{94.20±0.38} & \textbf{74.11±0.25} & \textbf{74.02±0.33} & \textbf{48.73±0.19} & \textbf{48.72±0.14} \\
    \bottomrule
    \end{tabular}
    }
\end{table}

\subsection{Ablation Studies}
\textbf{Effect of Reader and Proposer.} 
The Reader agent analyzes recent academic papers on neural architectures and related advances, and proposes key innovations and knowledge that could potentially improve neural network performance. 
Then the Reflector agent retrieves appropriate knowledge for the network to enhance it.
To validate the effectiveness of LLM-empowered Reader and LLM-empowered Proposer, we conducted ablation experiments, and the results are shown in Table ~\ref{tab:reader}.
we built two strong baselines.
When both Reader and Proposer are absent, we adopt a set of expert designed modification suggestions used in LeMo-NADe~\cite{rahman2024lemo} to generate design suggestions.
When only the Proposer is present, we prompts LLM to propose design suggestions based on its own knowledge without accessing to the recent achievements. 
When only the Reader is present, we randomly select knowledge from the knowledge pool as design suggestions.
We observe that both Reader and Proposer are important for improving the performance of NAD, especially on the challenging dataset, \ie ImageNet16-120. 

\begin{table}[t]
    \centering
    \caption{Ablation studies of the Developer Team. Experiments are conducted on our NAD benchmark. ``LIF'' and ``LDE'' mean learning from immediate feedback and learning from design experience respectively.
    }
    \scalebox{0.85}{
    \begin{tabular}{c|ccc|c|cc|c}
    \toprule
    Task & Graph & LIF & LDE & \# Tokens (K) & E & Q & SR \\
    \midrule
    \parbox{2.5mm}{\multirow{4}{*}{\rotatebox[origin=c]{90}{\textbf{Macro}}}}  
    & \xmark & \xmark & \xmark  & 2.23±0.27 & 0.64 & 0.63 & 0.40 \\
    & \cmark & \xmark & \xmark  &  0.58±0.66 & 0.54 & 0.78 & 0.42 \\
    & \cmark & \cmark & \xmark  &  0.73±0.82 & 0.64 & 0.84 & 0.54 \\
    & \cmark & \cmark & \cmark  & \textbf{0.53±0.68} & \textbf{0.78} & \textbf{0.87} & \textbf{0.68} \\
    \midrule
    \parbox{2.5mm}{\multirow{4}{*}{\rotatebox[origin=c]{90}{\textbf{Micro}}}}  
    & \xmark & \xmark & \xmark & 2.10±0.22 & 0.76 & 0.65 & 0.49 \\
    & \cmark & \xmark & \xmark &  0.49±0.63 & 0.62 & 0.97 & 0.60 \\
    & \cmark & \cmark & \xmark & 0.48±0.65 & 0.70 & 0.89 & 0.62 \\
    & \cmark & \cmark & \cmark & \textbf{0.31±0.45} & \textbf{0.92} & \textbf{0.96} & \textbf{0.88} \\
    \bottomrule
    \end{tabular}
    }
    \label{tab:develop_team}
\end{table}

\textbf{Effect of Graph-based Neural Architecture Representation.} We compare the graph-based neural architecture representation with the code-generation baseline. Specifically, we carefully design the prompts to prompt the LLM to directly generate codes for modifying the neural network architecture. From the experiments, we find that while existing LLMs demonstrate impressive capabilities in generating executable program codes (64\% executability), the quality is relatively low (63\% quality score). By adopting the graph representation, the LLMs can focus more on designing network architecture without distracted by the code style and formatting. In addition, the graph-based representation is more compact, resulting in fewer required tokens.

\textbf{Effect of Learning from Immediate Feedback (LIF).} 
The Reflector agent plays a crucial role in improving the executability and the quality of the modified network architecture. Adding LIF improves the executability from 0.54 to 0.64, while improving the quality score from 0.78 to 0.84.  

\textbf{Effect of Learning from Design Experience (LDE).}
The integration of learning from design experience significantly enhances the performance of NAD, in terms of both the executability and the quality, resulting in improved success rate.
For micro-level modifications, the code execution rate of the generated neural network reached 92\% and the success rate is significantly improved from 0.62 to 0.88.

\subsection{Large-scale NAD Experiment}
\begin{figure}[t]
    \centering
    \includegraphics[width=0.8\linewidth]{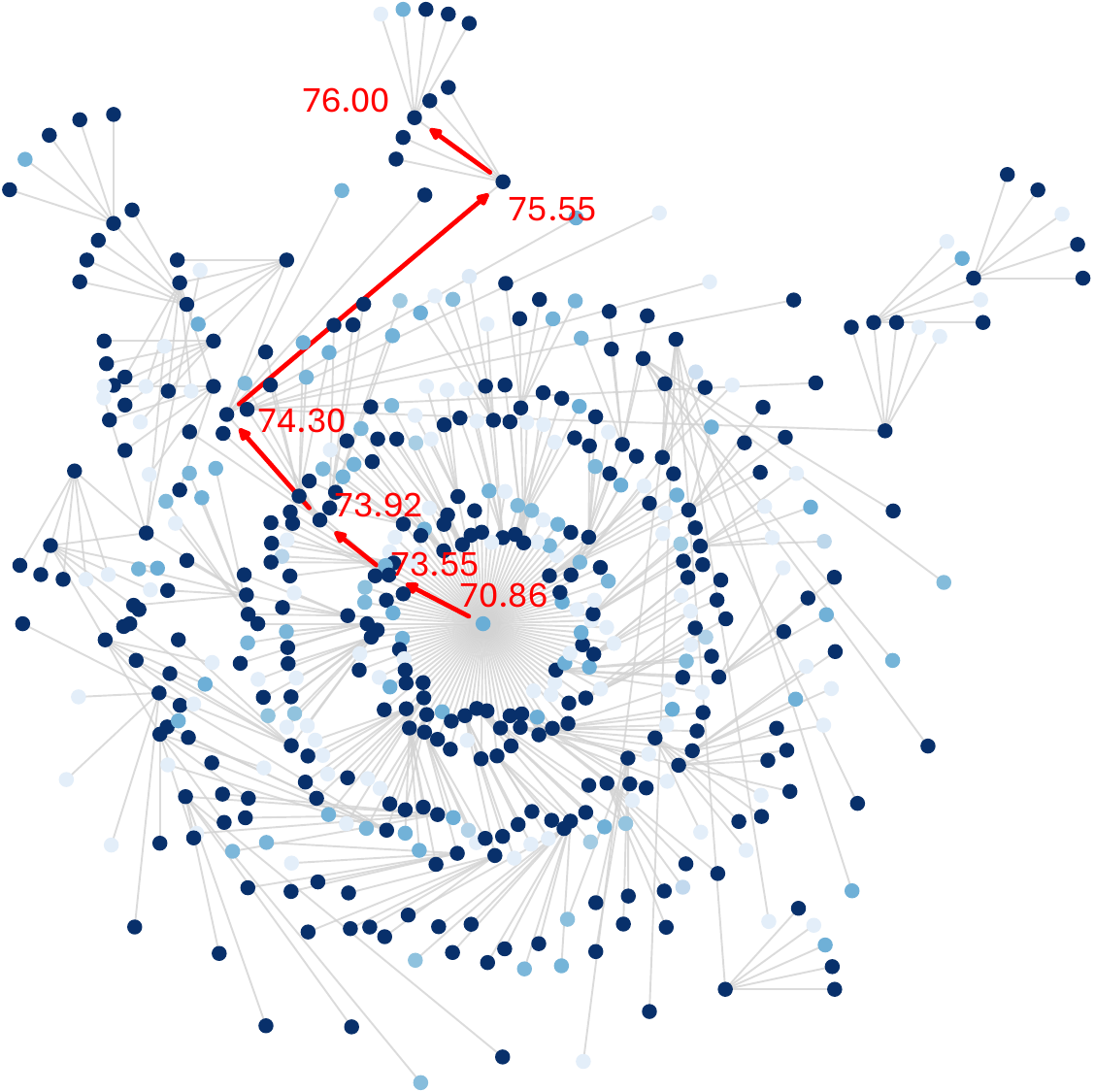}
    \caption{Large-scale NAD experiments.
    Each node represents a model, and the edge indicates which model is modified based on. The darker the node color, the higher the accuracy on the test set. 
    The root node in the center is ResNet.
    The path highlighted by the red arrows is the improvement path of the optimal model found.}
    \label{fig:large-scale}
\end{figure}
To further explore the potential of NADER, we use NADER to conduct a larger-scale neural architecture design on CIFAR100.
We initialize with ResNet~\cite{he2016deep} and generate 500 novel architectures using NADER. 
The modification process of these 500 architectures is shown in Figure~\ref{fig:large-scale}. 
To obtain these 500 architectures, we used 5,371K input tokens and 987K output tokens, totaling \$23 (\$2.50 per 1M input tokens and \$10.00 per 1M output tokens), resulting in an average cost of \$0.046 per architecture.
Finally, we get a novel architecture with an accuracy of 76.0\% on the test dataset, which is 5.14\% higher than the initial architecture. Futhermore, we evalute this architecture on the CIFAR-10 and ImageNet16-120. The experimental results, shown in Table ~\ref{tab:nasbench}, demonstrate that the designed architecture achieves strong performance across different datasets.

%% file: sections/5_conclusion.tex
\section{Conclusion}
In this paper, we propose NADER (Neural Architecture Design via multi-Agent collaboRation), a novel framework for neural architecture design (NAD). The framework adopts a multi-agent collaboration strategy to automatically explore novel neural architectures beyond predetermined search spaces. 
We propose Reader to continuously learn from recent academic literature, Proposer to raise design proposals, Modifier to implement these proposals, and Reflector to learn from feedback and historical records. Experiments demonstrate the effectiveness of the proposed method across a range of benchmark datasets.

%% file: sections/appendix.tex
\section{Detailed Experimental Setting}
\subsection{Macro Skeleton}
When conducting experiments on CIFAR10, CIFAR100~\cite{krizhevsky2009learning}, and ImageNet16-120~\cite{chrabaszcz2017downsampled}, in order to compare fairly with the compared methods, we constrain the model designed by our methods to have the same macro skeleton as NAS-Bench-201~\cite{dong2020bench}.
The skeleton is initiated with a stem block. The main body of the skeleton includes three stacks of cells, connected by a downsampling block.
Each cell has the same architecture and is stacked five times.
Each downsampling block downsamples the input feature map by two times.
Unlike NAS-Bench-201, we do not set a fixed model width.
When a new architecture is designed, we search for a suitable width for this model to constrain the number of parameters and FLOPs of the model so that it does not exceed the maximum values in NAS-Bench-201.
\subsection{Implement of Agents}
All our agents are implemented using GPT-4o.
And we use `text-embedding-ada-002' to extract text embeddings and build vector databases.
\begin{table}[ht]
    \centering
    \caption{The training hyperparameters for CIFAR-10 and CIFAR-100.}
    \scalebox{0.9}{
    \begin{tabular}{c|c||c|c}
    \toprule
    optimizer & SGD & initial LR & 0.1 \\
    Nesterov & \cmark & ending LR & 0 \\
    momentum & 0.9 & LR schedule & cosine \\
    weight decay & 0.0005 & epoch & 200 \\
    batch size & 256 & normalization & \cmark \\
    random flip & p=0.5 & random crop & size=32, padding=4 \\
    \bottomrule
    \end{tabular}\label{tab:hp_cifar}
    }
    \vspace{-5mm}
\end{table}
\subsection{Model Training Details}
When training and testing the model, we use the same training dataset, validation dataset, test dataset and training hyperparameters as NAS-Bench-201.
Table~\ref{tab:hp_cifar} shows the training hyperparameters for CIFAR-10 and CIFAR-100.
ImageNet16-120 has the same training hyperparameters as CIFAR-10 except that the random crop size is 16 and the padding is 2
We repeat the experiment three times using 777, 888 and 999 as the random number seeds.

\section{Prompt design and Output Examples for Agents}
\subsection{Reader}
The agent Reader has two LLM-based actions: selecting relevant papers and extracting knowledge from the relevant papers.
Tables \ref{tab:prompt_reader1} and \ref{tab:prompt_reader2} present the prompt templates for these two actions, respectively.
Table \ref{tab:example_insp} shows several examples of knowledge extracted from the papers.
\subsection{Proposer}
Table \ref{tab:prompt_proposer} presents the prompt template used by the agent Proposer to select modification suggestions.
\subsection{Modifier}
Table \ref{tab:prompt_dag} presents the prompt used to define the graph-based neural architecture representation. Table \ref{tab:prompt_modifier1} presents the prompt template and examples used by the Modifier to generate new architectures through multi-turn dialogues. In the first round, the Modifier generates a architecture containing an undefined operation. In the second round, after providing a prompt to the Modifier, it successfully generates a new architecture.
\begin{table}[htbp]
    \centering
    \caption{Prompt template for selecting relevant papers.}
    \resizebox{0.49\textwidth}{!}{
        \begin{tabular}{p{0.49\textwidth}}
        \hline
        \#\#\#Instruction\#\#\#\\
        You are a computer vision research specialist with a deep background in the field of computer vision, particularly in deep learning models and visual recognition tasks.\\
        You needs to evaluate a given paper to determine if it can inspire to design the better basic blocks architecture of vision models' backbone.\\
        \#\#\#Goal\#\#\#\\
        According to the title and abstract of the paper, analyzing whether can get inspiration from the paper to design the better basic architecture of visual model backbone.\\
        \#\#\#Constraints\#\#\#\\
        The inspiration must be related to the basic block architecture design of the visual model backbone.\\
        \#\#\#Workflow\#\#\#\\
        1. Read and understand the title and abstract of the paper.\\
        2. Summarizing the innovation and contribution of this paper.\\
        3. Analyzing whether you can get inspiration from this paper to design a better basic block architecture for visual models.\\
        \#\#\#Title\#\#\#\\
        \{\{\textit{title}\}\}\\
        \#\#\#Abstract\#\#\#\\
        \{\{\textit{abstract}\}\}\\
        \#\#\#Output\#\#\# \\
        Answer yes or no prefix with \#\#response\#\# in the end.\\
        \hline
        \end{tabular}}
    \label{tab:prompt_reader1}
\end{table}
\begin{table*}[htbp]
    \centering
    \caption{Prompt template for extracting knowledge from papers.}
    \begin{tabular}{p{0.96\textwidth}}
        \hline
        \#\#\#Instruction\#\#\#\\
        You are a computer vision research expert.
        Please list the inspirations you get from this paper to design the basic block architecture of the visual model backbone.\\
        \#\#\#Constraints\#\#\#\\
        1. A paper usually contains several sections: abstract, introduction, related work, methods, experiments and conclusion. Please focus on the methods of the paper to respond.\\
        2. Inspirations must to be detailed and related to designing the basic block architecture of the visual model backbone.\\
        \#\#\#Input\#\#\#\\
        The following is the content of the paper:\\
        \{\{\textit{paper}\}\}\\
        \#\#\#Output\#\#\#\\
        You response should wrap each inspirations with $<$inspiration$>$ and $<$/inspiration$>$, and use `,' to separate different knowledge.\\
        \hline
    \end{tabular}
    \label{tab:prompt_reader2}
\end{table*}
\begin{table*}[htbp]
    \centering
    \caption{Example of knowledge extracted from papers.}
    \begin{tabular}{l|p{0.86\textwidth}}
        \hline
        PaperId & Knowledge \\
        \hline
        603~\cite{Che_2023_CVPR} &
        \begin{list}{\arabic{enumi}.}{
            \usecounter{enumi}
            \setlength{\leftmargin}{8pt}
            \setlength{\itemsep}{0pt}
            \setlength{\topsep}{0pt}
            \setlength{\parsep}{0pt}
            \setlength{\partopsep}{0pt}
        }
            \item Design the Task Net to leverage knowledge co-embedding features constructed from both image quality and disease diagnosis, using multiple branches to specifically address different aspects of the input data.
            \item Implement a Global Attention Block within the Task Net to extract task-specific features, focusing on both channel and spatial attention to capture relevant features for each task.
        \end{list}\\
        \hline
        1530~\cite{Wen_2023_CVPR} &
        \begin{list}{\arabic{enumi}.}{
            \usecounter{enumi}
            \setlength{\leftmargin}{8pt}
            \setlength{\itemsep}{0pt}
            \setlength{\topsep}{0pt}
            \setlength{\parsep}{0pt}
            \setlength{\partopsep}{0pt}
        }
            \item Adoption of a multi-scale architecture using reversible residual blocks and squeeze modules to capture a wide range of style details while minimizing spatial information loss.
            \item Exclusion of normalization layers in the reversible blocks to facilitate learning direct style representation without interference, enhancing the style transfer fidelity.
            \item Implementing channel refinement in reversible residual blocks to manage redundant information accumulation and enhance stylization quality.
        \end{list}\\
        \hline
    \end{tabular}
    \label{tab:example_insp}
\end{table*}
\begin{table*}[htbp]
    \centering
    \caption{Prompt template for selecting modification suggestions.}
    \begin{tabular}{p{0.96\textwidth}}
        \hline
        \#\#\#Instruction\#\#\#\\
        You are a computer vision research expert, and you have deep insights into neural arcgitecture design.\\
        You will be given a block to be improved and several candidate inspirations, you need to compare the candidate inspirations and rank them according their usefulness for guiding the improvement of the block.\\
        \#\#\#Block\#\#\#\\
        The following is the block to be improved and candidate inspirations. The neural architecture of the block of the model is described in the form of a computational graph.\\
        \{\{\textit{block}\}\}\\
        \#\#\#Candidate inspirations\#\#\#\\
        The following are the candidate inspirations. Each inspiration is given in the form of `inspiration index:inspiration'.\\
        \{\{\textit{candidate inspirations}\}\}\\
        \#\#\#Output\#\#\#\\
        Please rank the all candidate inspirations in descending order according to their usefulness. You response should wrap all inspiration index of the inspirations with $<$response$>$ and $<$/response$>$, and use `,' to separate different indexes.\\
        \hline
    \end{tabular}
    \label{tab:prompt_proposer}
\end{table*}

\begin{table*}[htbp]
    \centering
    \caption{Prompt for defining the graph-based neural architecture representation.}
    \resizebox{0.99\textwidth}{!}{
    \begin{tabular}{p{0.99\textwidth}}
        \hline
        Each block starts with ``\#\#block\_name\#\#''. In each line, you can use the ``index:operation'' to define the node of computation graph or use the ``index1-$>$index'' to define the edge of computation graph.\\
        The following is a list of available operations:\\
        \vspace{-5mm}
        \begin{list}{}{\setlength{\leftmargin}{8mm}}
            \setlength{\itemsep}{0pt}
            \setlength{\parskip}{0pt}
            \item \textbf{Conv2d(out\_channels, kernel\_size, stride, dilation, groups)} Two-dimensional convolution operation, `out\_channels' represents the output dimension; `kernel\_size' represents the convolution kernel size; `stride' represents the step size, default: 1; `dilation' is the hole convolution size, default: 1; `groups' groups number of the channels, default:1.
            \item \textbf{Linear(out\_channels)} Linear fully connected layer, `out\_channels' represents the output dimension.
            \item \textbf{AvgPool2d(kernel\_size, stride)} Two-dimensional average pooling operation, `kernel\_size' represents the kernel size, `stride' represents the step size.
            \item \textbf{MaxPool2d(kernel\_size,stride)} Two-dimensional maximum pooling operation, `kernel\_size' represents the kernel size, `stride' represents the step size.
            \item \textbf{AdaptiveMaxPool2d(output\_size)} Two-dimensional maximum pooling operation pools the input feature map into a feature map with a length and width of output\_size. For example, AdaptiveMaxPool2d(output\_size=1) pools a feature map of the shape of (B,C,H,W) into (B,C,1,1) shape.
            \item \textbf{AdaptiveAvgPool2d(output\_size)} Two-dimensional average pooling operation.
            \item \textbf{Add} Tensor-by-element addition operation, the input tensors' shape must conform to the broadcasting rule.
            \item \textbf{Mul} Tensor-by-element multiplication operation, the input tensors' shape must conform to the broadcasting rule.
            \item \textbf{Multiply} Matrix multiplication operation, the entered tensor shapes must conform to the tensor multiplication rule.
            \item \textbf{concat(dim)} Tensor concating operation, all tensors input to this operation are concated in the dim dimension. The sizes of the concated tensors dimensions other than the dim dimension should be consistent. For example, concat(dim=1) concates all input tensors in the 1 dimension.
            \item \textbf{mean(dim)} Average the tensor in dim dimension. For example, mean(dim=1) pools a input tensor of shape (B,L,D) into the output tensor of shape (B,1,D) by average in the dimension 1.
            \item \textbf{max(dim)} Maximize the tensor in dim dimension. For example, max(dim=2) pools a input tensor of shape (B,L,D) into the output tensor of shape (B,L,2) by max in the dimension 2.
            \item \textbf{sum(dim)} Sum the tensor in dim dimension. For example, sum(dim=0) pools a input tensor of shape (B,L,D) into the output tensor of shape (0,L,D) by sum in the dimension 0.
            \item \textbf{softmax(dim)} Apply a softmax operation at dim dimension. For example, softmax(dim=1) calculate the softmax of input tensor with shape (B,L,D) and the output tensor's shape is (B,L,D).
        \end{list}
        \vspace{-3mm}
        The activation functions that can be used are: \textbf{ReLU}, \textbf{GELU}, \textbf{Sigmoid}.\\
        The normalization methods that can be used are:\\
        \vspace{-5mm}
        \begin{list}{}{\setlength{\leftmargin}{8mm}}
            \setlength{\itemsep}{0pt}
            \setlength{\parskip}{0pt}
            \item \textbf{BN} Batch normalization
            \item \textbf{LN} Layer normalization.
        \end{list}
        \vspace{-3mm}
        The tensor can be transformed by using the following operations:\\
        \vspace{-5mm}
        \begin{list}{}{\setlength{\leftmargin}{8mm}}
            \setlength{\itemsep}{0pt}
            \setlength{\parskip}{0pt}
            \item \textbf{permute(*dims)} rearranges the tensor dimensions, `dims' is the order of the new dimensions, for example: permute(0, 2, 3, 1) changes the tensor shape from (B, C, H, W) to (B, H, W, C).
            \item \textbf{repeat(*sizes)} repeats the tensor along the specified dimensions. `sizes' is a list containing the number of repetitions along each dimension. For example: repeat(1, 3, 2, 4) repeats the tensor 1 times in the first dimension, 3 times in the second dimension, 2 times in the third dimension, 4 times in the forth dimension.
            \item \textbf{reshape(*shape)} changes the shape of the tensor to the specified shape; `shape' is an array representing the shape of new tensor; you can use -1 as the size of a dimension to automatically calculate the size of the dimension to ensure that the total number of elements remains unchanged; for example: reshape(B, H, W, C) means changing the shape of the tensor to (B, H, W, C).
        \end{list}
        \vspace{-3mm}
        Variables you may use include:\\
        \vspace{-5mm}
        \begin{list}{}{\setlength{\leftmargin}{8mm}}
            \setlength{\itemsep}{0pt}
            \setlength{\parskip}{0pt}
            \item \textbf{input}: input feature map, the shape is (B,C,H,W).
            \item \textbf{output}: output feature map, the shape is (B,dim,H,W). Noting that the output node can have only one input.
            \item \textbf{C}: the number of channels of the input feature map.
            \item \textbf{dim}: the number of channels of the output feature map.
            \item \textbf{H}: the height of the input feature map.
            \item \textbf{W}: the width of the input feature map.
        \end{list}
        \vspace{-3mm}
        You can use the basic \textbf{$+$},\textbf{$-$},\textbf{$\times$},\textbf{$/$} operations.\\
        \hline
    \end{tabular}
    }
    \label{tab:prompt_dag}
\end{table*}
\begin{table*}[hbtp]
    \centering
    \caption{Prompt template and example for generating new architecture.}
    \resizebox{\textwidth}{!}{
    \begin{tabular}{l|l}
        \hline
        Human & \makecell[l]{
        \#\#\#Instruction\#\#\#\\
        You are an expert who is proficient in various model structures of deep learning.\\
        Please make reasonable modifications to the specified block based on the characteristics of the block and the inspiration.\\
        \#\#\#Constraints\#\#\#\\
        1. Please ensure that the number of input channels and output channels of the generated block are both C.\\
        2. Note that structures in the modified block that unrelated to the proposal should be kept as original as possible.\\
        3. The new neural architecture you output must comply with the BlockDefinition format.\\
        \#\#\#BlockDefinition\#\#\#\\
        \{\{\textit{Definition of the graph-based neural architecture representation}\}\}\\
        \#\#\#block\#\#\#\\
        \{\{\textit{block}\}\}\\
        \#\#\#inspiration\#\#\#\\
        \{\{\textit{inspiration}\}\}\\
        \#\#\#Design Experience\#\#\#\\
        Refer to the following suggestions to help you generate a block that better meets block definition.\\
        \{\{\textit{Experience in generating the correct architectures}\}\}\\
        Refer to the following suggestions to help you generate a block that has better performance.\\
        \{\{\textit{Experience in generating better-performing architectures}\}\}\\
        \#\#\#Output\#\#\#\\
        When outputting, you only need to output the block that meet the defined rules, and do not output other irrelevant information.\\}\\
        \hline
        Assistant & \makecell[l]{\includegraphics[width=0.96\textwidth]{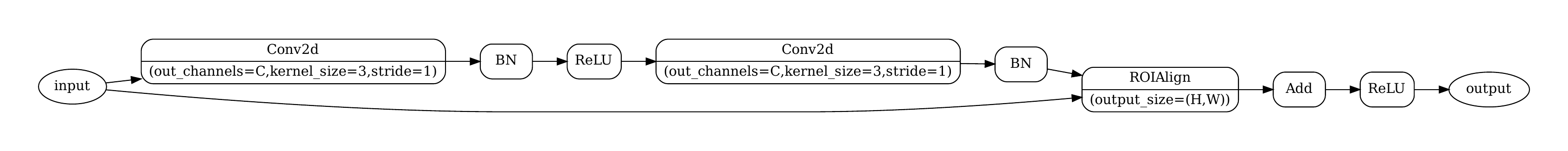}} \\
        \hline
        Reflector & \{`status':`error',`context':`node 8 error: Undefined computation ROIAlign is used'\}\\
        \hline
        Human & \makecell[l]{The block you generate has following error: \{`status':`error',`context':`node 8 error: Undefined computation ROIAlign is\\ used'\}, please fix it and generate a new one.}\\
        \hline
        Assistant & \makecell[l]{\includegraphics[width=0.96\textwidth]{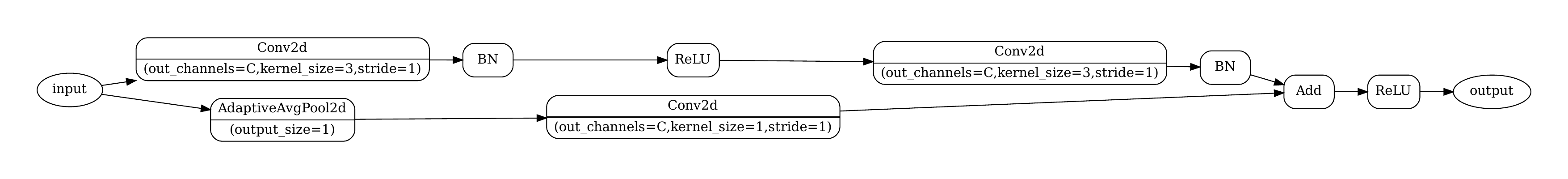}} \\
        \hline
        Reflector & \{`status':`success'\}\\
        \hline
    \end{tabular}
    }
    \label{tab:prompt_modifier1}
\end{table*}
\subsection{Reflector}
Reflector reflects on the historical records of faulty architectures and performance-degrading architectures generated by the Modifier and generating experience in generating correct neural architectures and experience in designing performance-enhanced architectures, respectively. Table \ref{tab:prompt_reflector1} presents the prompt template for reflecting on historical records of faulty architectures and Table \ref{tab:prompt_reflector2} presents the prompt template for reflecting on historical records of performance-degrading architectures. Tables \ref{tab:example_exp1} and Table \ref{tab:example_exp2} present several examples of the two types of experience, respectively.
\begin{table*}[t]
    \centering
    \caption{Prompt template for reflecting on the historical records of faulty architectures.}
    \resizebox{0.99\textwidth}{!}{
    \begin{tabular}{p{0.99\textwidth}}
        \hline
        \#\#\#Instruction\#\#\#\\
        You are an expert who is proficient in neural architecture design. 
        The structure of neural networks is now described in terms of directed acyclic graphs. The following is the definition of directed acyclic graphs.\\
       \{\{\textit{Definition of the graph-based neural architecture representation}\}\}\\
        \#\#\#Input\#\#\#\\
        Now there is a network that does not fully meet the above definition and a hint of reason:\\
        \{\{\textit{block}\}\}\\
        Error reason: \{\{\textit{error}\}\}\\
        \#\#\#output\#\#\#\\
        Please analyze the reason of network design errors, and based on this, give a general design tip to prompt users to accurately design a network that fully meets the requirements. The tip should be wrapped in $<$tip$>$ and $<$/tip$>$.\\
        \hline
    \end{tabular}
    }
    \label{tab:prompt_reflector1}
\end{table*}
\clearpage
\begin{table*}[htbp]
    \centering
    \caption{Prompt template for reflecting on the historical records performance-degrading architectures.}
    \resizebox{0.99\textwidth}{!}{
    \begin{tabular}{p{0.99\textwidth}}
        \hline
        \#\#\#Instruction\#\#\#\\
        You are an expert who is proficient in neural architecture design.
        You will be given a raw model and a modified model and their corresponding accuracy on test dataset and you need to analyze why the accuracy of the raw model decreases after modification, and give a suggestion to avoid this error in the next modification.\\
        The structure of neural networks is now described in terms of directed acyclic graphs. The following is the definition of directed acyclic graphs:\\
        \{\{\textit{Definition of the graph-based neural architecture representation}\}\}\\
        \#\#\#Input\#\#\#\\
        Their is the raw model and its accuracy:\\
        \{\{\textit{raw block}\}\}\\
        Accuracy: \{\{\textit{raw block's accuracy\}}\}\\
        Their is the modified model and its accuracy:\\
        \{\{\textit{new block}\}\}\\
        Accuracy: \{\{\textit{new block's accuracy}\}\}\\
        \#\#\#Constrain\#\#\#\\
        1. The suggestion must be relevant to the neural network structure.\\
        2. The suggestion must be a sentence no more than 50 words.\\
        3. The suggestion must be must be general.\\
        \#\#\#Output\#\#\#\\
        Please think step by step about the reasons why the accuracy of the model decreases after modification and give a suggestion to avoid this error in the next modification.
        The suggestion should be wrapped in $<$suggestion$>$ and $<$/suggestion$>$.\\
        \hline
    \end{tabular}
    }
    \label{tab:prompt_reflector2}
\end{table*}
\begin{table*}[t]
    \centering
    \caption{Example of experience in generating correct archs.}
    \resizebox{0.99\textwidth}{!}{
    \begin{tabular}{p{0.99\textwidth}}
        \hline
        \begin{list}{\arabic{enumi}.}{
            \usecounter{enumi}
            \setlength{\leftmargin}{8pt}
            \setlength{\itemsep}{0pt}
            \setlength{\topsep}{0pt}
            \setlength{\parsep}{0pt}
            \setlength{\partopsep}{0pt}
        }
            \item When using LN, always ensure that the last dimension of the input tensor is the channel dimension. You can achieve this by using the `permute' operation to rearrange the tensor dimensions appropriately before applying LN.
            \item When designing a neural network, ensure that the `out\_channels' parameter in the `Conv2d' operation is divisible by the `groups' parameter. This is necessary to maintain valid configurations and avoid errors. Specifically, if you set `groups` to be equal to `C' (the number of input channels), make sure that `out\_channels' is also a multiple of `C'.
            \item When designing a neural network with multiple branches that will be concatenated or added together, ensure that all branches maintain consistent spatial dimensions (height and width) throughout their respective operations. Use padding in convolution and pooling layers to preserve the input dimensions, or carefully adjust stride and kernel sizes to ensure outputs are compatible for concatenation or addition.
        \end{list}\\
        \hline
    \end{tabular}
    }
    \label{tab:example_exp1}
\end{table*}
\begin{table*}[htbp]
    \centering
    \caption{Example of experience in generating correct archs.}
    \resizebox{0.99\textwidth}{!}{
    \begin{tabular}{p{0.99\textwidth}}
        \hline
        \begin{list}{\arabic{enumi}.}{
            \usecounter{enumi}
            \setlength{\leftmargin}{8pt}
            \setlength{\itemsep}{0pt}
            \setlength{\topsep}{0pt}
            \setlength{\parsep}{0pt}
            \setlength{\partopsep}{0pt}
        }
            \item Ensure modifications retain the original block's dimensional consistency and spatial information to maintain performance.
            \item Ensure that added layers and operations contribute to meaningful feature extraction and avoid unnecessary complexity.
            \item Consider maintaining intermediate bottleneck layers and incorporating attention or pooling mechanisms to enhance feature extraction.
            \item Ensure that any added skip connections or operations do not disrupt the learning process by excessively altering the network's expected data flow.
            \item Ensure the newly added paths or operations do not interfere destructively with existing paths, and validate their impact on gradient flow.
        \end{list}\\
        \hline
    \end{tabular}
    }
    \label{tab:example_exp2}
\end{table*}
\clearpage

\section{Detailed Experimental Results}
\subsection{Detailed Numberical Results}

\begin{table}[htbp]
    \centering
    \caption{Detailed Experimental Results on CIFAR-10 with ResNet initialization.}
    \scalebox{0.99}{
    \begin{tabular}{c|cc|cc}
    \toprule
    \multirow{2}{*}{Trail}  & \multicolumn{2}{c|}{5 archs} & \multicolumn{2}{c}{10 archs} \\
                           & validation & test & validation & test \\
    \midrule
    Trail 1 & 91.28 & 94.83 & 91.28 & 94.83 \\
    Trail 2 & 90.83 & 94.37 & 90.86 & 94.37 \\
    Trail 3 & 91.39 & 94.35 & 91.39 & 94.35 \\
    \bottomrule
    \end{tabular}\label{tab:detail_resnet_cifar10}}
    \vspace{0.5cm}
    \caption{Detailed Experimental Results on CIFAR-100 with ResNet initialization.}
    \vspace{-0.2cm}
    \scalebox{0.99}{
    \begin{tabular}{c|cc|cc}
    \toprule
    \multirow{2}{*}{Trail}  & \multicolumn{2}{c|}{5 archs} & \multicolumn{2}{c}{10 archs} \\
                           & validation & test & validation & test \\
    \midrule
    Trail 1 & 74.18 & 74.18 & 75.11 & 75.11 \\
    Trail 2 & 73.84 & 73.57 & 74.34 & 74.45 \\
    Trail 3 & 71.64 & 71.62 & 74.44 & 74.38 \\
    \bottomrule
    \end{tabular}\label{tab:detail_resnet_cifar100}}
    \vspace{0.5cm}
    \caption{Detailed Experimental Results on ImageNet16-120 with ResNet initialization.}
    \vspace{-0.2cm}
    \scalebox{0.99}{
    \begin{tabular}{c|cc|cc}
    \toprule
    \multirow{2}{*}{Trail}  & \multicolumn{2}{c|}{5 archs} & \multicolumn{2}{c}{10 archs} \\
                           & validation & test & validation & test \\
    \midrule
    Trail 1 & 47.53 & 47.72 & 49.27 & 49.58 \\
    Trail 2 & 49.00 & 48.52 & 49.00 & 48.52 \\
    Trail 3 & 47.40 & 47.72 & 47.40 & 47.72 \\
    \bottomrule
    \end{tabular}\label{tab:detail_resnet_imagenet16}}
    \vspace{0.5cm}
    \caption{Detailed Experimental Results on CIFAR-10 with Random initialization.}
    \vspace{-0.2cm}
    \scalebox{0.99}{
    \begin{tabular}{c|cc|cc}
    \toprule
    \multirow{2}{*}{Trail}  & \multicolumn{2}{c|}{5 archs} & \multicolumn{2}{c}{10 archs} \\
                           & validation & test & validation & test \\
    \midrule
    Trail 1 & 90.39 & 93.67 & 90.96 & 94.13 \\
    Trail 2 & 90.83 & 94.37 & 90.86 & 94.37 \\
    Trail 3 & 91.50 & 94.55 & 91.66 & 94.69 \\
    \bottomrule
    \end{tabular}\label{tab:detail_random_cifar10}}
    \vspace{0.5cm}
    \caption{Detailed Experimental Results on CIFAR-100 with Random initialization.}
    \vspace{-0.2cm}
    \scalebox{0.99}{
    \begin{tabular}{c|cc|cc}
    \toprule
    \multirow{2}{*}{Trail}  & \multicolumn{2}{c|}{5 archs} & \multicolumn{2}{c}{10 archs} \\
                           & validation & test & validation & test \\
    \midrule
    Trail 1 & 74.04 & 74.35 & 74.04 & 74.35 \\
    Trail 2 & 73.84 & 73.57 & 74.34 & 74.45 \\
    Trail 3 & 74.44 & 74.15 & 74.86 & 74.72 \\
    \bottomrule
    \end{tabular}\label{tab:detail_random_cifar100}}
    \vspace{0.5cm}
    \caption{Detailed Experimental Results on ImageNet16-120 with Random initialization.}
    \vspace{-0.2cm}
    \scalebox{0.99}{
    \begin{tabular}{c|cc|cc}
    \toprule
    \multirow{2}{*}{Trail}  & \multicolumn{2}{c|}{5 archs} & \multicolumn{2}{c}{10 archs} \\
                           & validation & test & validation & test \\
    \midrule
    Trail 1 & 48.60 & 48.82 & 50.63 & 50.38 \\
    Trail 2 & 49.00 & 48.52 & 49.00 & 48.52 \\
    Trail 3 & 48.60 & 48.82 & 50.57 & 50.00 \\
    \bottomrule
    \end{tabular}\label{tab:detail_random_imagenet16}}
\end{table}
Table~\ref{tab:detail_resnet_cifar10}, Table~\ref{tab:detail_resnet_cifar100}, and Table~\ref{tab:detail_resnet_imagenet16} show the experimental results on CIFAR10, CIFAR100, and ImageNet16-120 using ResNet as the initial model.
Table~\ref{tab:detail_random_cifar10}, Table~\ref{tab:detail_random_cifar100}, and Table~\ref{tab:detail_random_imagenet16} show the experimental results on CIFAR10, CIFAR100, and ImageNet16-120 using random models in NAS-Bench-201 as the initial models.
We repeat each experiment with different random seeds for three times.

\subsection{Analysis of Large-scale NAD Experiment Results}
\begin{figure}[htbp]
    \centering
    \includegraphics[width=0.96\linewidth]{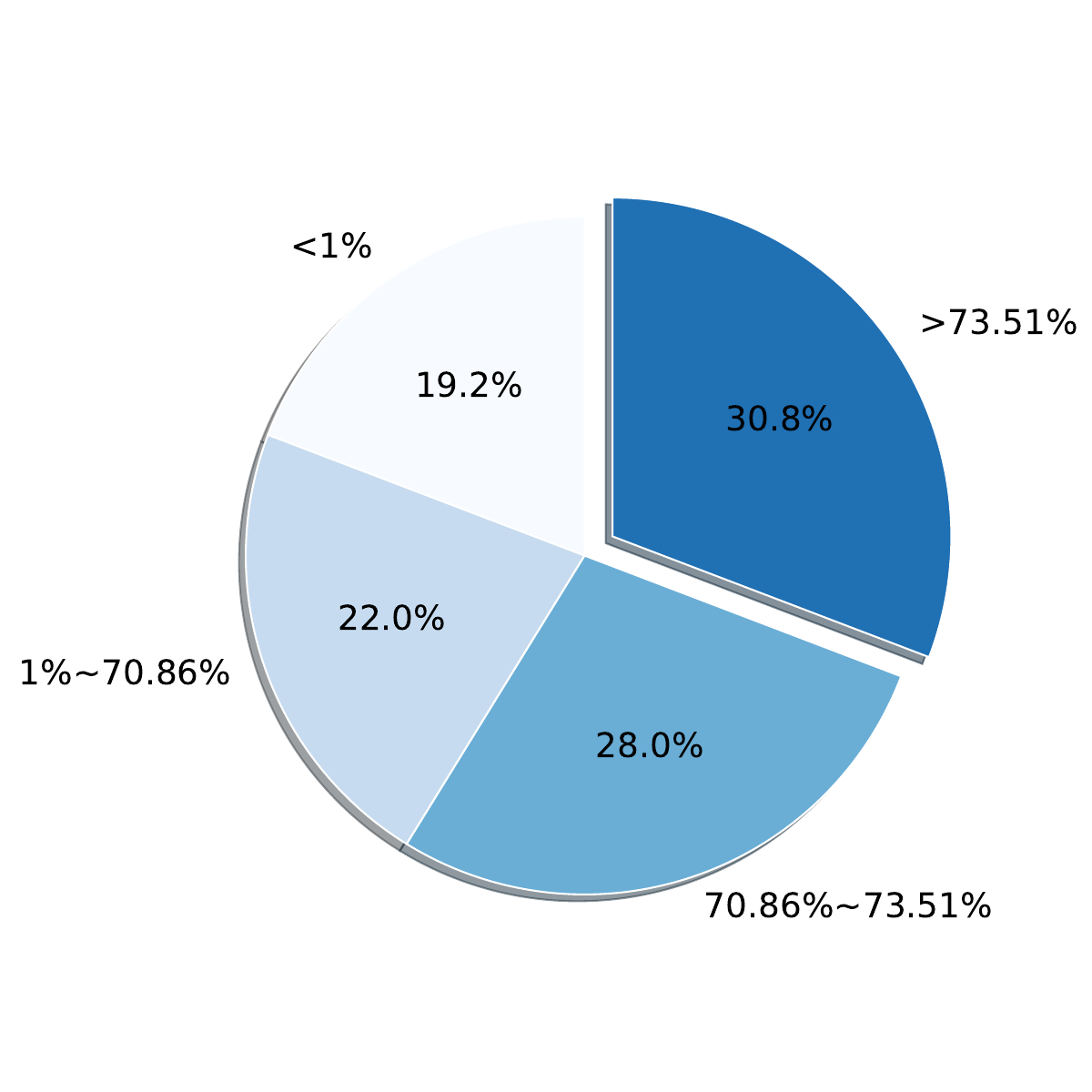}
    \caption{Distribution of test accuracy of 500 models designed by NADER on CIFAR-100.}
    \label{fig:pie}
\end{figure}

Figure~\ref{fig:pie} illustrates the distribution of test accuracy of 500 models designed by NADER on CIFAR-100.
It can be seen that 30.8\% of the models have an accuracy that exceeds the optimal model in the neural architecture search space designed by NAS-Bench-201.
58.8\% of the proposals generated by the Proposer are effective, and the performance of the model based on these proposals surpasses that of the initial model.
19.2\% of the models failed to converge during training due to unreasonable design.
At the same time, these bad models exhibit noticeable differences in loss values within the initial few epochs, allowing for early detection and termination to reduce unnecessary training.
\subsection{Generalization Ability of the Model Designed by NADER}
After 500 designs, NADER designed a model on the CIFAR-100 dataset that showed significant improvements compared to the optimal model in the NAS method’s search space (NB201-Optimal). We refer to this model as NADER-500.
We further retrain and test the NADER-500 on several other datasets to evaluate the generalization ability of the model designed by NADER. The experimental results are presented in Table \ref{tab:nader-500}. It show that NADER-500 demonstrates significant advantages on datasets involving natural scenes, such as animals and vehicles (e.g., CIFAR-100~\cite{krizhevsky2009learning}, STL-10~\cite{2011An}, GT SRB~\cite{2012Man}). 

\begin{table}[th]
    \caption{Experimental results of NADER-500 and NB201-Optimal on different datasets. The test accuracy is presented.}
    \begin{tabular}{ccc}
    \hline
    Dataset   & \begin{tabular}[c]{@{}c@{}}NB201-Optimal\\ (\#Params:1.30M)\end{tabular} & \begin{tabular}[c]{@{}c@{}}NADER-500\\ (\#Params:1.28M)\end{tabular} \\ \hline
    CIFAR100~\cite{krizhevsky2009learning}  & 73.23 & 75.97 \\
    GTSRB~\cite{2012Man}    & 96.16 & 96.94 \\
    STL10~\cite{2011An}     & 69.65  & 72.13 \\
    \hline
    \end{tabular}
    \label{tab:nader-500}
\end{table}

\subsection{Architecture Details}
\begin{figure}[htbp]
    \centering
    \includegraphics[width=0.6\linewidth]{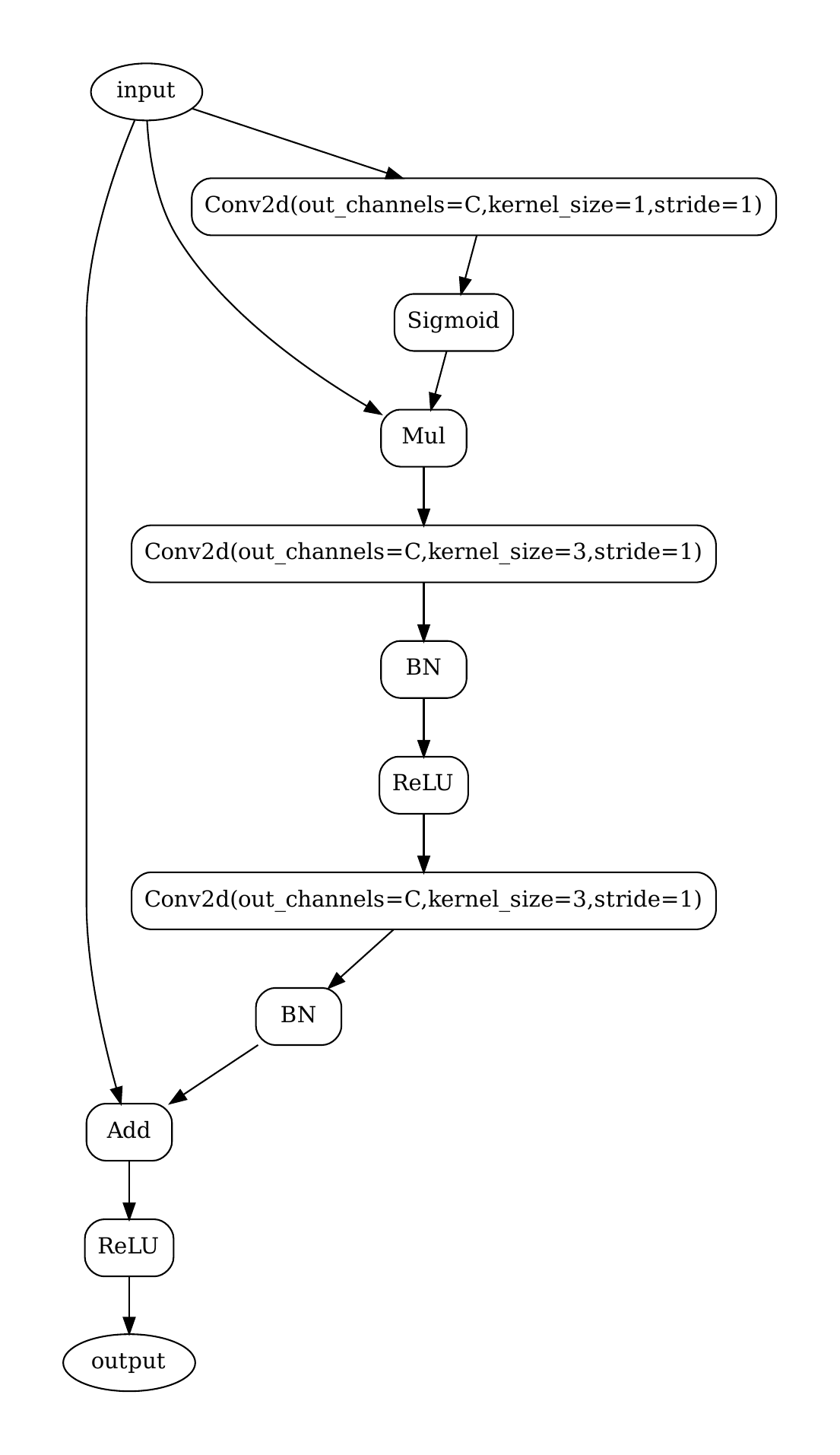}
    \caption{The neural architecture of the optimal model obtained by performing 10 iterations of search on CIFAR-10. The accuracy on the test dataset is 94.83\%.}
    \label{fig:dag_cifar10}
\end{figure}

\begin{figure}[htbp]
    \centering
    \includegraphics[width=0.9\linewidth]{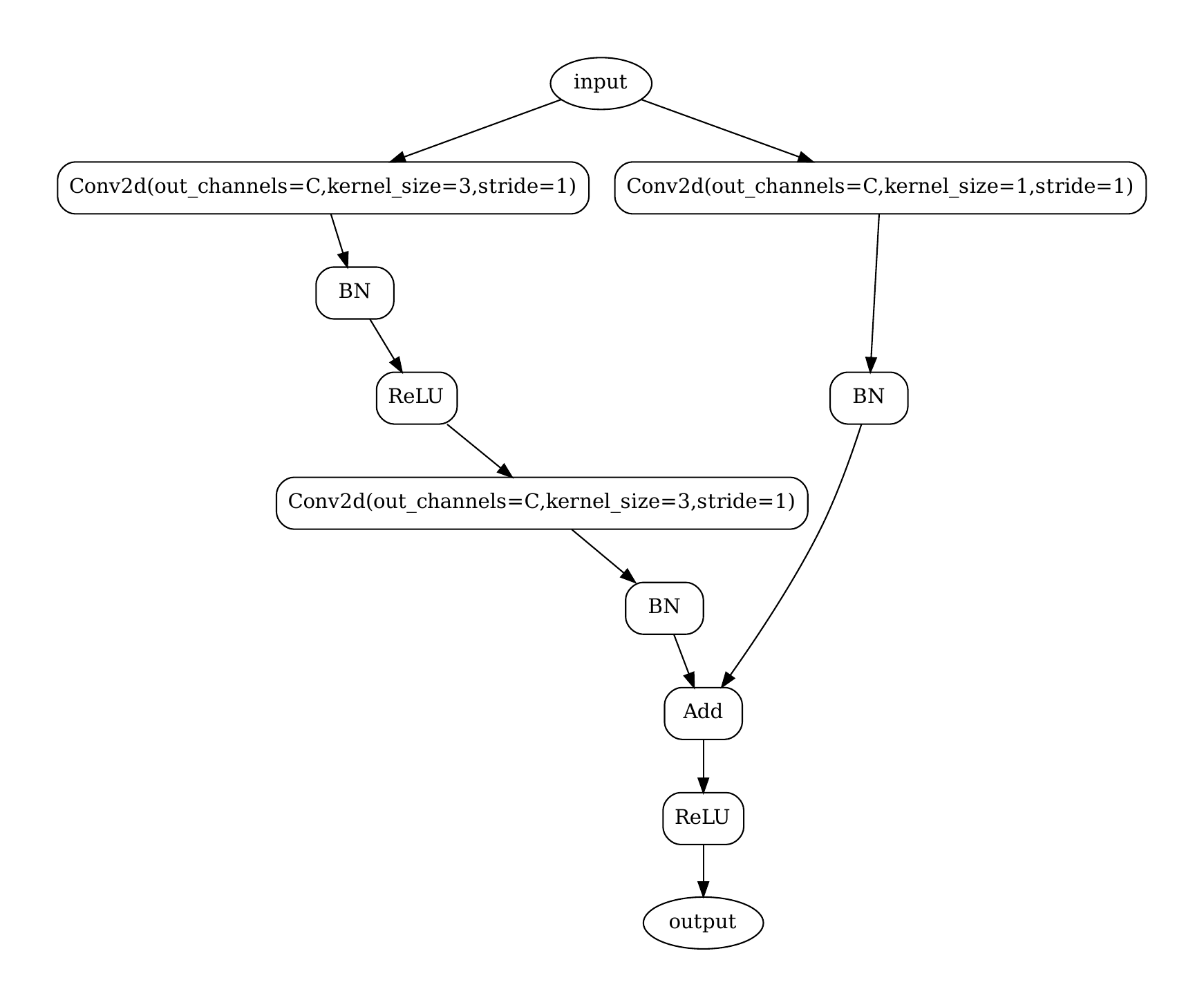}
    \caption{The neural architecture of the optimal model obtained by performing 10 iterations of search on ImageNet16-120. The accuracy on the test dataset is 49.58\%.}
    \label{fig:dag-imagenet16}
\end{figure}

\begin{figure}[h]
    \centering
    \includegraphics[width=0.9\linewidth]{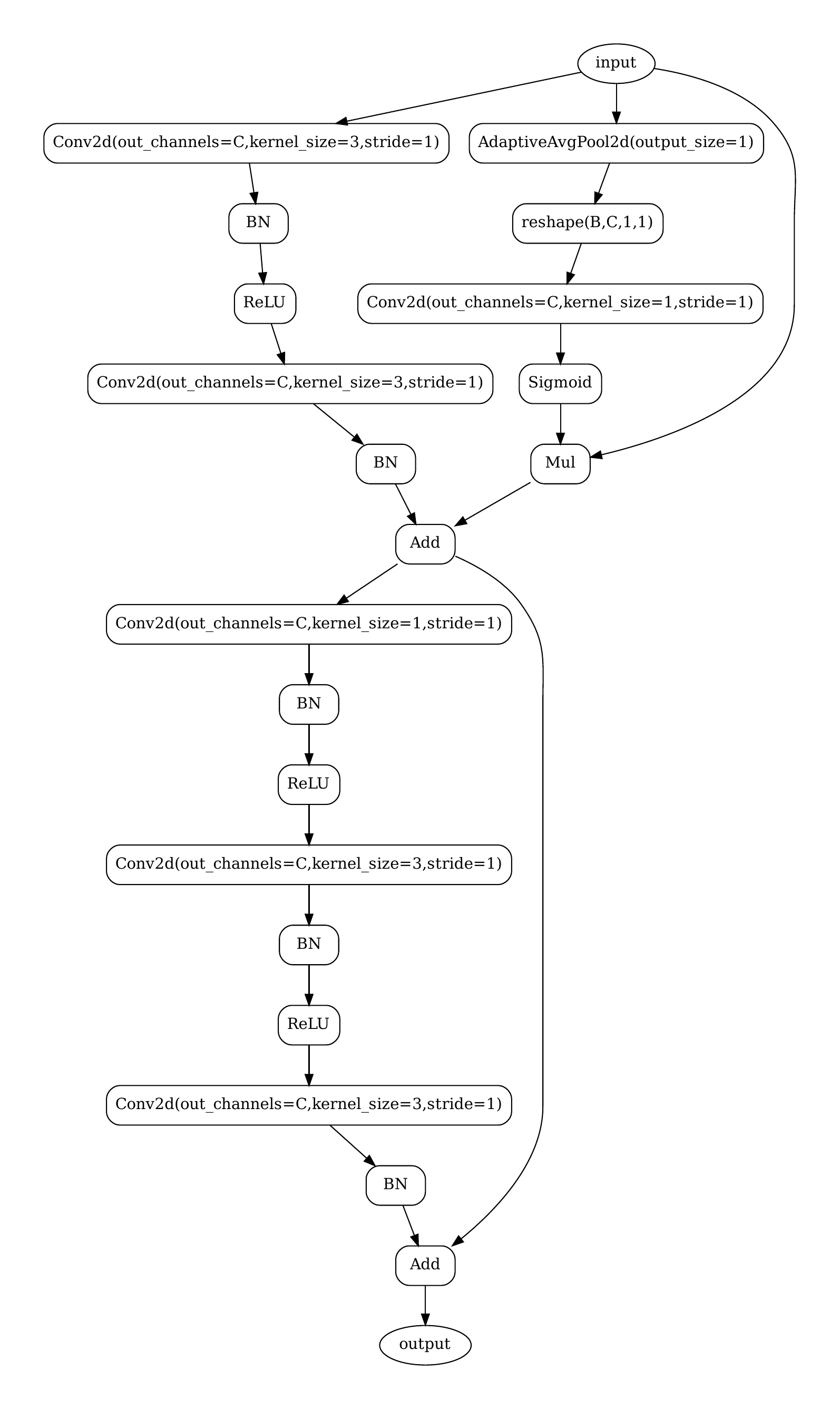}
    \caption{The neural architecture of the optimal model obtained by performing 10 iterations of search on CIFAR-100. The accuracy on the test dataset is 75.11\%.}
    \label{fig:dag_cifar100}
\end{figure}

\begin{figure}[htbp]
    \centering
    \includegraphics[width=0.96\linewidth]{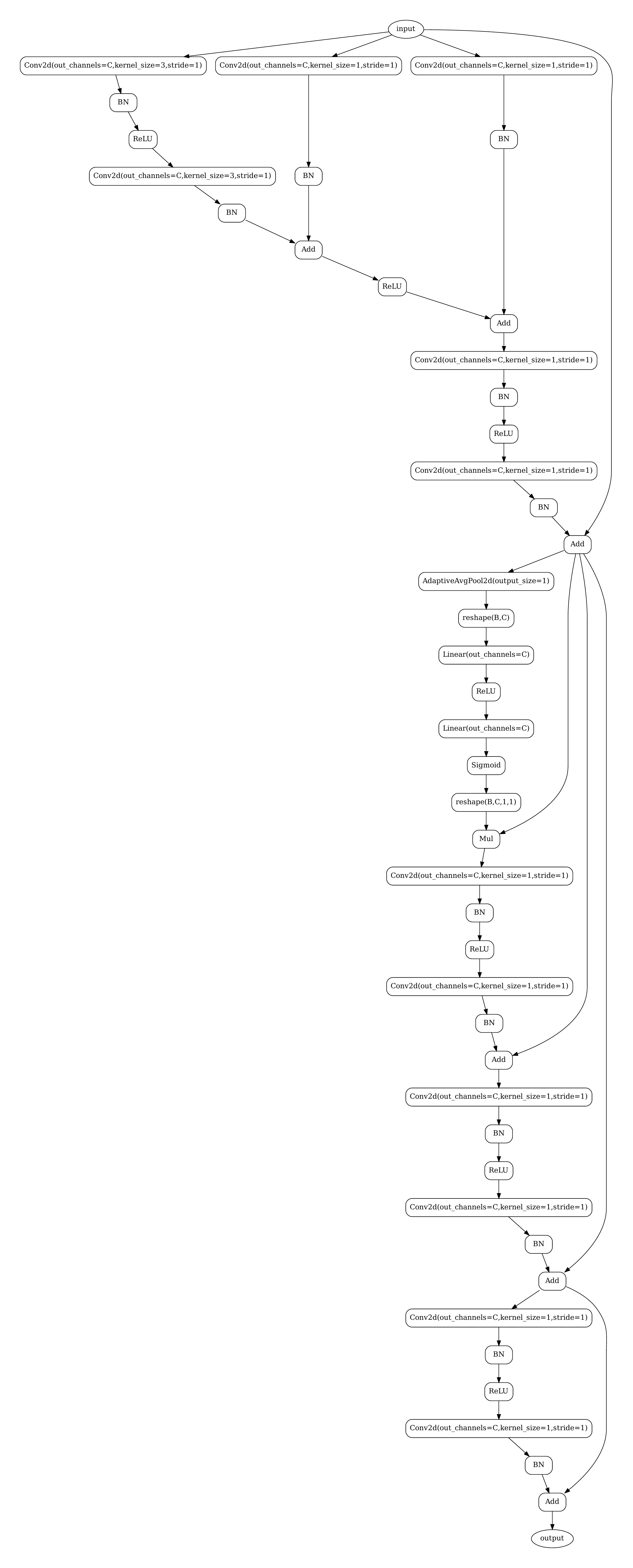}
    \caption{The neural architecture of the optimal model obtained by performing 500 iterations of search on CIFAR-100. The accuracy on the test dataset is 76.00\%.}
    \label{fig:exa_dag76}
\end{figure}

Figure~\ref{fig:dag_cifar10}, Figure~\ref{fig:dag_cifar100}, Figure~\ref{fig:dag-imagenet16} and Figure~\ref{fig:exa_dag76} show several examples of neural architectures designed by NADER with ResNet as the initial model.
It can be found that the multi-agent collaboration method we designed can stimulate the creativity of LLMs and generate novel and effective architectures, rather than simply generate architectures that LLMs may have seen in the training process.
\clearpage
